\documentclass[sigconf,9pt]{acmart}

\renewcommand\footnotetextcopyrightpermission[1]{}

\usepackage{subcaption}
\usepackage[ruled,linesnumbered]{algorithm2e}
\usepackage{tikz}
\usepackage{pifont}
\usepackage{multirow}
\usepackage{dblfloatfix}
\usepackage[graphicx]{realboxes}
\usepackage{titlesec}
\usepackage{comment}
\usepackage[para]{threeparttable}
\usepackage{amsmath}

\titleformat{\paragraph}
{\normalfont\normalsize\bfseries}{\theparagraph}{1em}{}
\titlespacing*{\paragraph}
{0pt}{3.25ex plus 1ex minus .2ex}{1.5ex plus .2ex}

\renewcommand{\vec}[1]{\mathbf{#1}} 

\newcommand{\sect}{\textsection}

\begin{document}
\title{Indoor Smartphone SLAM with Learned Echoic Location Features}


\author{Wenjie Luo}
\affiliation{%
  \department{Singtel Cognitive and AI Lab for Enterprises}
  \institution{Nanyang Technological University}
  \country{Singapore}
}
\authornote{Also with School of Computer Science and Engineering, Nanyang Technological University, Singapore.}

\author{Qun Song}
\affiliation{%
  \department{ERI@N, Interdisciplinary Graduate School}
  \institution{Nanyang Technological University}
  \country{Singapore}
}
\authornotemark[1]

\author{Zhenyu Yan}
\affiliation{%
  \department{Department of Information Engineering}
  \institution{The Chinese University of Hong Kong}
  \country{HongKong}
}

\author{Rui Tan}
\affiliation{%
  \department{Singtel Cognitive and AI Lab for Enterprises}
  \institution{Nanyang Technological University}
  \country{Singapore}
}
\authornotemark[1]

\author{Guosheng Lin}
\affiliation{%
  \department{School of Computer Science and Engineering}
  \institution{Nanyang Technological University}
  \country{Singapore}
}

\renewcommand{\shortauthors}{}

\begin{abstract}
Indoor self-localization is a highly demanded system function for smartphones. The current solutions based on inertial, radio frequency, and geomagnetic sensing may have degraded performance when their limiting factors take effect. In this paper, we present a new indoor simultaneous localization and mapping (SLAM) system that utilizes the smartphone's built-in audio hardware and inertial measurement unit (IMU). Our system uses a smartphone's loudspeaker to emit near-inaudible chirps and then the microphone to record the acoustic echoes from the indoor environment. Our profiling measurements show that the echoes carry location information with sub-meter granularity. To enable SLAM, we apply contrastive learning to construct an echoic location feature (ELF) extractor, such that the loop closures on the smartphone's trajectory can be accurately detected from the associated ELF trace. The detection results effectively regulate the IMU-based trajectory reconstruction. Extensive experiments show that our ELF-based SLAM achieves median localization errors of $0.1\,\text{m}$, $0.53\,\text{m}$, and $0.4\,\text{m}$ on the reconstructed trajectories in a living room, an office, and a shopping mall, and outperforms the Wi-Fi and geomagnetic SLAM systems.
\end{abstract}


    


\maketitle
\pagestyle{plain} 

\section{Introduction}
\label{sec:intro}



Location sensing is a fundamental service of mobile operating systems. As of May 2022, 73\% of the top 100 free Android apps on Google Play \cite{top-free} require location information. In the last two decades, research has exploited various smartphone's built-in sensing modalities for indoor location sensing, which include Wi-Fi \cite{bahl2000radar,youssef2005horus}, GSM \cite{hightower2005learning}, FM radio \cite{chen2012fm}, visible light \cite{zhang2016litell}, imaging \cite{gao2015sextant}, acoustic background \cite{tarzia2011indoor}, and geomagnetism \cite{he2017geomagnetism}. However, these sensing modalities have their own limiting factors. For instance, radio frequency (RF) signals are susceptible to electromagnetic noises and transceivers' automatic gain controls. Visible light sensing suffers blockage. Visual imaging may generate privacy concerns in certain spaces and times. The acoustic background only gives room-level granularity. Hence, identifying new modalities from smartphones' common built-in hardware to enrich the location sensing toolbox has been an interest of research.

Exploiting smartphone's built-in audio hardware for active location sensing receives increasing research \cite{rossi2013roomsense,tung2015echotag,song2018deep,pradhan2018smartphone,zhou2017batmapper,lian2021echospot}. The active sensing uses smartphone's loudspeaker to emit excitation sounds in the target indoor space and microphone to capture the echoes from the surroundings. The echoes carry information that depicts how a sound emitted from a specific location reverberates in the indoor space. The existing approaches can be classified into two categories. The first category, namely, {\em analytic approach} \cite{lian2021echospot,pradhan2018smartphone,zhou2017batmapper}, estimates the smartphone's location by analyzing the processes of the sound reflections by nearby surfaces (e.g., walls). The image source modeling (ISM) is a prevailing analysis technique. However, when the surroundings are complex (e.g., irregular surfaces, many nearby objects with complex 3D structures, etc), accurate ISM becomes intractable. Thus, the existing studies following the analytic approach often make simplifying assumptions that the major reflectors are at most two nearby walls \cite{lian2021echospot,pradhan2018smartphone,zhou2017batmapper}. The second category, namely, {\em fingerprint approach} \cite{rossi2013roomsense,song2018deep,tung2015echotag}, uses the echoes captured by the smartphone at a certain location as the fingerprint of the location and then applies supervised learning to build a location recognition model. However, the blanket process of collecting labeled fingerprints at spatially fine-grained locations to form the training dataset imposes a high overhead. Thus, the existing studies only target room-level localization \cite{rossi2013roomsense,song2018deep} or recognize a limited number of locations (11 closed locations in \cite{tung2015echotag}).


Nevertheless, the fingerprint approach has the potential in offering good generalizability because it does not make specific assumptions about the surroundings. Moreover, the high spatial resolution achieved in \cite{tung2015echotag} encourages further investigation of whether satisfactory resolutions can be maintained when the number of fingerprinted locations increases. To develop a preliminary understanding, in this paper, we conduct a blanket process of fingerprinting 128 locations using excitation chirps with near-inaudible frequencies in a $16 \times 28\,\text{m}^2$ space hosting tens of cubicles and several meeting areas. Our analysis by applying supervised learning to the collected data shows that the fingerprint approach can maintain sub-meter localization accuracy. The results suggest that the acoustic echo is a promising modality for designing useful indoor localization services for off-the-shelf smartphones.

To unleash the fingerprint approach from the blanket labeled training data collection process, in this paper, we aim to design a simultaneous localization and mapping (SLAM) system based on the smartphone's inertial measurement unit (IMU) data and the acoustic echoes. Specifically, when the user carrying the smartphone moves in the space, both IMU data and acoustic echo data are collected. When the user returns to a location visited earlier, the movement trajectory produces a {\em loop closure}. If the loop closures can be accurately detected using the echo data, the IMU-based dead reckoning result, which is deviation-prone, can be rectified to yield a more accurate trajectory reconstruction. Accordingly, the reconstructed trajectory and the associated echo fingerprints form a {\em trajectory map}. The trajectory map only contains the echoes from the space covered by the trajectory. The extension to the whole indoor space relies on the {\em map superimposition} of many users' trajectory maps.

As a key to SLAM, loop closure detection requires an effective embedding such that any two acoustic echo samples captured at the same location are close in the embedded space, while those collected at different locations are apart. 
However, finding an effective embedding by which a certain similarity metric can effectively signal loop closures is challenging. Our tests show that the generic acoustic features, e.g., power spectral density, spectrogram, and principal component analysis, cannot effectively discriminate locations. Thus, we resort to learning an effective embedding. A straw-man proposal is to train a deep neural network (DNN) using labeled echo data collected at sparse locations and use the output of a hidden layer as the embedding. However, such embedding for recognizing the pre-defined locations in general becomes ineffective for the locations out of the training dataset such as the arbitrary locations on the user's trajectory in SLAM. Contrastive learning (CL), on the other hand, exploits the self-supervised learning technique to learn effective representations from unlabeled data. Applied to our SLAM problem, it can learn from the unlabeled echo data collected on the user's trajectory and only requires the information of whether two echo samples are collected at close locations. Such coarse proximity information can be easily derived using heuristics. Thus, we apply CL with a cosine similarity-based contrastive loss function to train an embedding DNN that outputs a representation called {\em echoic location feature} (ELF). As such, the cosine similarity between any two echo samples' ELFs effectively signals whether the two samples are collected at the same/close locations (i.e., loop closure).

To realize the ELF-based SLAM, we make the following three designs. First, we design a trajectory-level CL procedure to learn the trajectory-specific ELFs for loop closure detection. It consists of the \textit{pre-training} step, which uses extensive synthetic echoes to build a basic ELF extractor, and the \textit{fine-tuning} step, which adapts the basic extractor to a target indoor space using unlabeled echoes collected on the user trajectory. Second, we design a clustering-based approach to curate the loop closures. This is because, although the similarity values among the ELFs can signal most of the true positive loop closures, they can also cause false positives. The curation exploits prior knowledge about the user's movement to effectively remove the false positives. Third, we design a floor-level CL procedure to superimpose the trajectory maps from many users to form a single \textit{floor map}. The procedure reconciles the differences among the ELFs from different trajectory maps at the same location due to the non-negligible impact of smartphone orientation on echo data.

The contributions of this paper are summarized as follows:
\begin{itemize}
\item We conduct extensive profiling measurements to investigate the acoustic echo's spatial distinctness under a supervised learning setting. Our results suggest a sub-meter spatial resolution limit and show that acoustic echo is a promising modality for indoor location sensing with sub-meter accuracy. 
\item We design ELF with CL to detect loop closures for SLAM. We further design ELF-SLAM, a smartphone SLAM system using IMU and ELF data. We also apply CL to reconcile and superimpose the maps from many users. 

\item We conduct extensive experiments in a living room, an office, and a shopping mall. ELF-SLAM achieves sub-meter mapping and localization accuracy and outperforms the Wi-Fi and geomagnetic SLAMs. We also study the allowable intensities and/or needed mitigation for various practical affecting factors, including nearby people, audible noises, space layout changes, and audio hardware heterogeneity. 
\end{itemize}


The rest of this paper is organized as follows. \sect\ref{sec:related} reviews related work. \sect\ref{sec:pre} presents preliminaries. \sect\ref{sec:measure} presents the measurement study. \sect\ref{sec:design} presents the design of ELF-SLAM. \sect\ref{sec:eval} presents the evaluation results. 
\sect\ref{sec:conclude} concludes this paper.


\section{Related Work}
\label{sec:related}

\begin{table*}
  \caption{Infrastructure-free acoustic indoor localization and mapping.}  
  \label{tab:chirp-design-acoustic-sensing}
  \vspace{-1em}
  \begin{threeparttable}
    \centering
    \small
  \begin{tabular}{c|c|c|c|c|c|c|c|c}
    \hline
    \multirow{2}{*}{{\bf Approach}} & \multirow{2}{*}{{\bf Study}} & \multirow{2}{*}{{\bf Objective}} & {\bf Presumption} & {\bf Labeled} & \multirow{2}{*}{{\bf Resolution}} &\multicolumn{3}{c}{\bf Sensing technique}\\ 
    \cline{7-9}
     &  &  & {\bf on reflectors} & {\bf training data} &  & {\bf Mode} & {\bf Sensing signal} & {\bf Duration}\\ 
    \hline
    \multirow{4}{*}{Analytic} & VoLoc \cite{shen2020voice} & \multirow{2}{*}{Localization} & \multirow{4}{*}{Yes} & \multirow{4}{*}{No} & \multirow{4}{*}{Decimeters} & Passive & Human voice & 15 cmds \\
    \cline{2-2} \cline{7-9}
     & EchoSpot \cite{lian2021echospot} &  &  &  &  & \multirow{3}{*}{Active} & 18-23kHz FMCW\tnote{1} & $0.2\,\text{s}$ \\
     \cline{2-3} \cline{8-9}
     & BatMapper \cite{zhou2017batmapper} & \multirow{2}{*}{Mapping} &  &  &  &  & 8-16kHz chirps & $0.04\,\text{s}$ \\
     \cline{2-2} \cline{8-9}
     & SAMS \cite{pradhan2018smartphone} &  &  &  &  &  & 11-21kHz FMCW & $0.03\,\text{s}$ \\
    \hline
    \multirow{7}{*}{Fingerprint} & SurroundSense \cite{azizyan2009surroundsense} & \multirow{4}{*}{Localization} & \multirow{7}{*}{No} & \multirow{5}{*}{Yes} & Semantic locations & \multirow{2}{*}{Passive} & Acoustic & $60\,\text{s}$ \\
    \cline{2-2} \cline{6-6} \cline{9-9}
     & Batphone \cite{tarzia2011indoor} &  &  &  & \multirow{3}{*}{(Sub-)room-level} &  & background & $10\,\text{s}$ \\
     \cline{2-2} \cline{7-9}
     & RoomSense \cite{rossi2013roomsense} &  &  &  &  & \multirow{5}{*}{Active} & 0-24kHz MLS\tnote{2} & $0.68\,\text{s}$ \\
     \cline{2-2} \cline{8-9}
     & DeepRoom \cite{song2018deep} &  &  &  &  &  & 20kHz tone & $0.1\,\text{s}$ \\
     \cline{2-3} \cline{6-6} \cline{8-9}
     & EchoTag \cite{tung2015echotag} & Location tagging &  &  & Centimeter &  & 11-22kHz chirp & $0.42\,\text{s}$ \\
     \cline{2-3} \cline{5-6} \cline{8-9}
     & \multirow{2}{*}{{\bf ELF-SLAM}} & Localization &  & \multirow{2}{*}{No} & \multirow{2}{*}{Sub-meter} &  & \multirow{2}{*}{15-20kHz chirp} & \multirow{2}{*}{$0.1\,\text{s}$} \\
     &  & \& mapping &  &  &  &  &  &  \\
    \hline
    \end{tabular}
    \begin{tablenotes}\footnotesize
        \item[1] Frequency-modulated continuous-wave
        \item[2] Maximum Length Sequence
    \end{tablenotes}
  \end{threeparttable}
\end{table*}

$\blacksquare$ {\bf Acoustics-based indoor localization and mapping:} The ubiquity of audio speakers and microphones on consumer electronics has attracted research interest in acoustics-based indoor localization. The infrastructure-based approaches deploy dedicated sound beacons or receivers to localize a mobile receiver or beacon. However, the overhead of deploying the infrastructure is undesirable. Our review focuses on infrastructure-free approaches, which are summarized in Table~\ref{tab:chirp-design-acoustic-sensing}. 
The {\em analytic approach} analyzes the acoustic signal propagation processes. It passively senses the sounds from the source or actively emits acoustic signals and analyzes the echoes. VoLoc \cite{shen2020voice} uses a smart speaker to localize a user based on the angle-of-arrival (AoA) of the user's voice and the wall reflection. In EchoSpot \cite{lian2021echospot}, a device emits near-inaudible acoustic signals and analyzes the times of flight of the signals reflected off the human body and a nearby wall for user spotting. These studies \cite{shen2020voice,lian2021echospot} require presumption on the sound reflectors for triangulation.


The analytic approach can be also applied for indoor mapping by estimating the distances of the smartphone to the nearby surfaces (e.g., sidewalls, ceilings, and floors) \cite{zhou2017batmapper,pradhan2018smartphone}. To build the wall contour map, the studies \cite{zhou2017batmapper,pradhan2018smartphone} require the user to walk straight along the walls. They detect up to two walls simultaneously. In this work, we aim to build the maps on arbitrary trajectories with loop closures and the complete floor maps in indoor spaces. In addition, we aim to be able to address complex spaces such as a lab with dense cubicles and a shopping mall with dense stores.
The work \cite{krekovic2016echoslam} considers collated omni-directional speaker and microphone mounted on a robot and analyzes echoes' arrival times for indoor mapping. The evaluation in \cite{krekovic2016echoslam} is only based on simulations. In real world, complex surroundings (e.g., many nearby objects) may bring challenges to the echo arrival identification. 

Vis-\`{a}-vis analytic approach, the {\em fingerprint approach} collects training samples from different indoor locations and trains a machine learning model for online location inference \cite{azizyan2009surroundsense,tarzia2011indoor,rossi2013roomsense,tung2015echotag,song2018deep}. Early studies \cite{azizyan2009surroundsense, tarzia2011indoor,rossi2013roomsense} apply shallow learning and require either long recording times that may cause privacy concerns or full-spectrum recording that is susceptible to interference like human voice \cite{song2018deep}. The work \cite{song2018deep} applies deep learning to reduce the requirements on recording time and spectrum usage. These studies \cite{azizyan2009surroundsense,tarzia2011indoor,rossi2013roomsense,song2018deep} address semantic or room-level localization. 
The work \cite{tung2015echotag} uses echoic fingerprints to tag up to 11 locations with centimeters resolution. When the fingerprint approach is applied for the whole indoor space, the blanket process of collecting labeled fingerprints at many locations causes high overhead. 

\begin{figure}
    \centering
    \begin{subfigure}{0.35\columnwidth}
      \centering
      \includegraphics[width=\columnwidth]{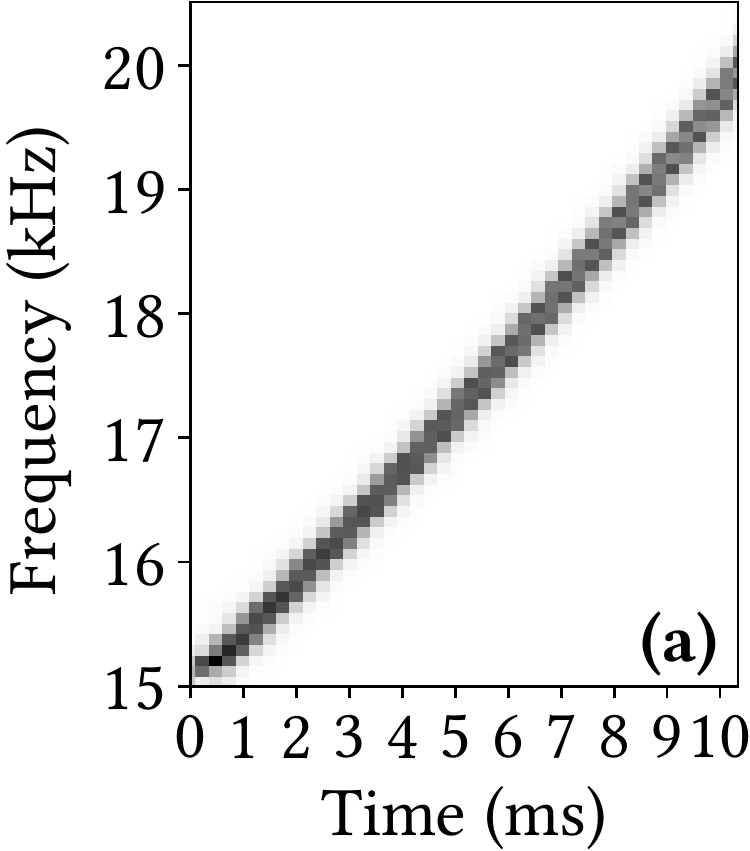} 
    \end{subfigure}
    \begin{subfigure}{0.6\columnwidth}
      \centering
      \includegraphics[width=\columnwidth]{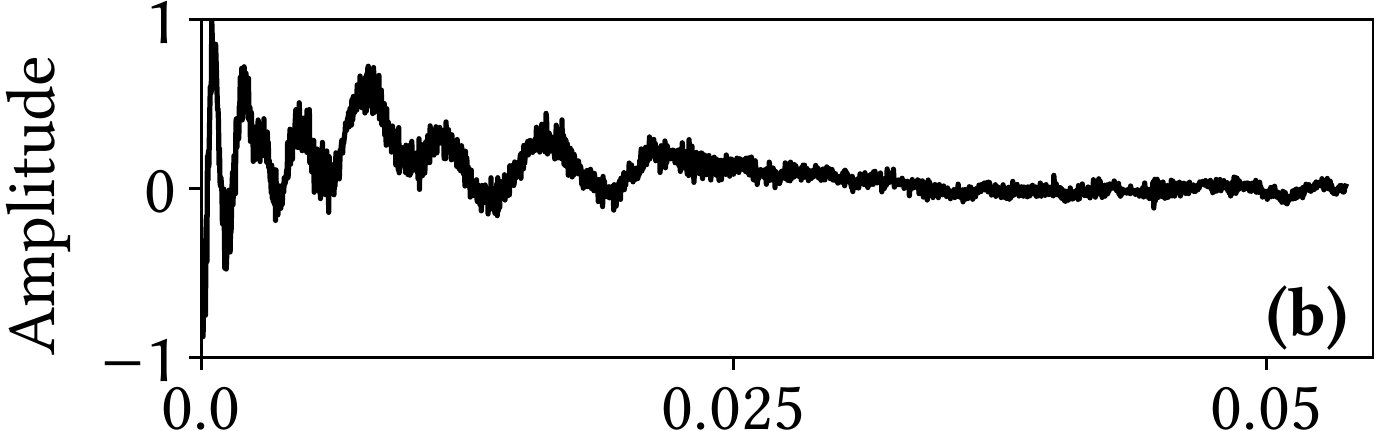} 
          \includegraphics[width=\columnwidth]{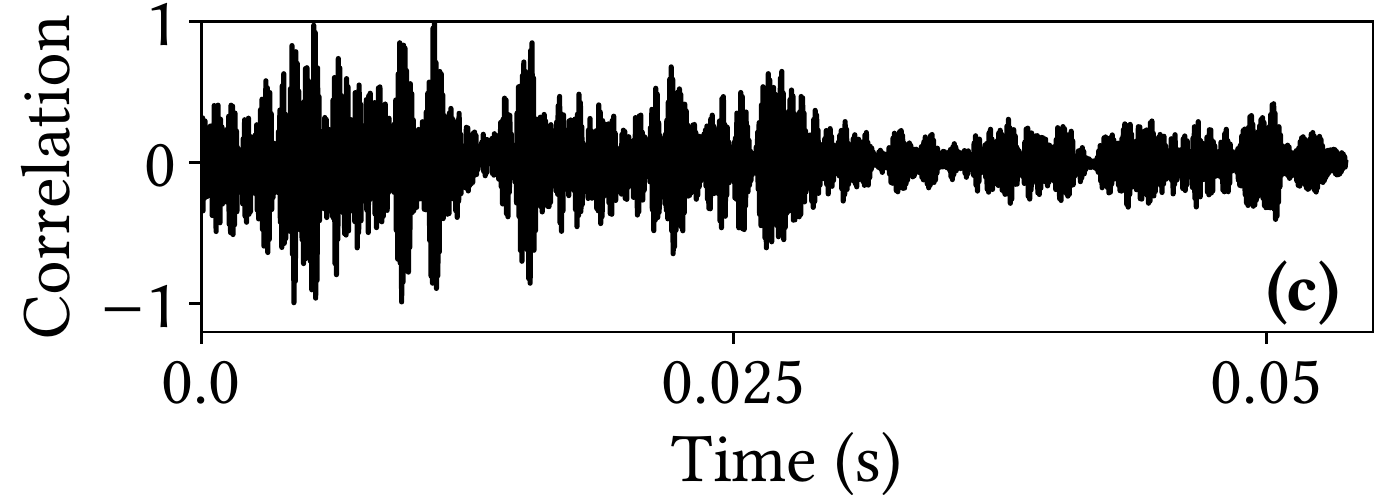}
    \end{subfigure}
    \vspace{-1em}
    \caption{(a) Excitation chirp's spectrogram; (b) Received signal in time domain; (c) Pearson correlation between received signal and excitation signal template.}
    \label{fig:signal}
    \vspace{-1em}
  \end{figure}
Acoustic sensing has also been used for other applications like inter-smartphone ranging \cite{peng2007beepbeep,zhang2012swordfight}, finger \cite{nandakumar2016fingerio,yun2017strata} and body \cite{nandakumar2017covertband} tracking, gesture recognition \cite{gupta2012soundwave}, speaker authentication \cite{chauhan2017breathprint}, breathing rate estimation \cite{xu2019breathlistener}, and lung function monitoring \cite{song2020spirosonic}.


$\blacksquare$ {\bf SLAM:} SLAM constructs the map of an environment in terms of a certain signal and localizes the user device simultaneously. Any SLAM solution consists of two components: {\em front-end signal processing} and {\em back-end pose-graph optimization}. 
Here, we discuss existing SLAM solutions according to their used sensing modalities and the used loop closure detection methods. Radar SLAM \cite{marck2013indoor} and Lidar SLAM \cite{droeschel2018efficient} are based on point clouds generated by radar and high-profile lidar, which are unavailable on smartphones.
Visual SLAM \cite{taketomi2017visual} uses the camera to capture images for landmark detection and map construction. The imaging may introduce privacy concerns. 
The imaging may introduce privacy concerns.  
Wi-Fi SLAM \cite{ferris2007wifi,arun2022p2slam} employs the received signal strength indicators (RSSIs) from nearby Wi-Fi access points. However, Wi-Fi RSSI can be time-varying. Geomagnetic SLAM \cite{wang2016keyframe} exploits the spatially varying magnetic field. Electromagnetic radiation (EMR) SLAM \cite{lu2018simultaneous} can use the smartphone's earphone as a side-channel sensor to sense the EMR from the alternating current power network. However, the side-channel sensing may experience weak signal strength when the earphone is away from the powerlines. This paper's evaluation will employ geomagnetic, EMR, and Wi-Fi SLAMs as the main baselines. 
Acoustic SLAM \cite{evers2018acoustic} constructs a map based on the AoAs from multiple infrastructural sound sources. Our ELF-SLAM is a new SLAM solution based on infrastructure-free acoustic echoes.

Different sensing modalities need specific loop closure detection methods. The DNN-based loop closure detection on Lidar and Visual SLAM \cite{zhang2021visual,cattaneo2022lcdnet} often outperforms the hand-crafted solutions. Wi-Fi SLAM \cite{ferris2007wifi} applies the Gaussian process latent variable model to determine the locations of the Wi-Fi RSSI signal. Both the geomagnetic and the EMR SLAM systems employ dynamic time warping (DTW) for loop closure detection. Different from these sensing modalities, generic acoustic features are ineffective to determine locations. Thus, we apply CL to learn a new acoustic feature embedding called ELF to effectively signal loop closures.

\section{Preliminaries}
\label{sec:pre}



$\blacksquare$ {\bf Excitation signal:}
The commonly used excitation signals include single tone, frequency hopping spectrum spread (FHSS), and chirp \cite{cai2022ubiquitous}. 
Single tone facilitates detecting Doppler shift and is usually adopted in Doppler ranging and tracking \cite{gupta2012soundwave}.
However, in this work, we prefer a multi-frequency excitation signal such that the echoes carry richer information about the surroundings for location fingerprinting. FHSS signal has rapidly hopping frequency in a wide band. As it better copes with self-interference, it is suitable for short-range active sensing \cite{yun2017strata}.
When exciting sizable indoor spaces, the self-interference can be easily avoided by limiting the excitation signal duration. Therefore, in this work, we adopt chirp with frequency varying with time continuously that can induce information-richer echoes, compared with FHSS.

As commodity smartphones support audio sampling at $44.1$ or $48\,\text{ksps}$, they can capture acoustic frequencies up to $22.05$ or $24\,\text{kHz}$. To reduce annoyance to human user, the excitation signal should be in the inaudible or near-inaudible frequency range.
Although nominal audible frequency is up to $20\,\text{kHz}$, the audible limit of average adults is usually $15$ to $17\,\text{kHz}$ \cite{audiblespectrum}.
Signals with frequencies higher than $20\,\text{kHz}$ often suffer from drastic distortions due to the smartphone audio hardware's nonlinearity in the inaudible band \cite{song2018deep}. In this paper, we adopt a near-inaudible logarithmic chirp sweeping the $15$--$20\,\text{kHz}$ band within $10\,\text{ms}$ as the excitation signal. Fig.~\ref{fig:signal}a shows the spectrogram of the excitation chirp. The chirp uses a relatively wide band (i.e., $5\,\text{kHz}$) for the benefit of pulse compression \cite{lazik2012indoor}, which helps capture fine-grained spatial features.



\begin{figure}
  \centering
  \includegraphics[width=0.8\columnwidth]{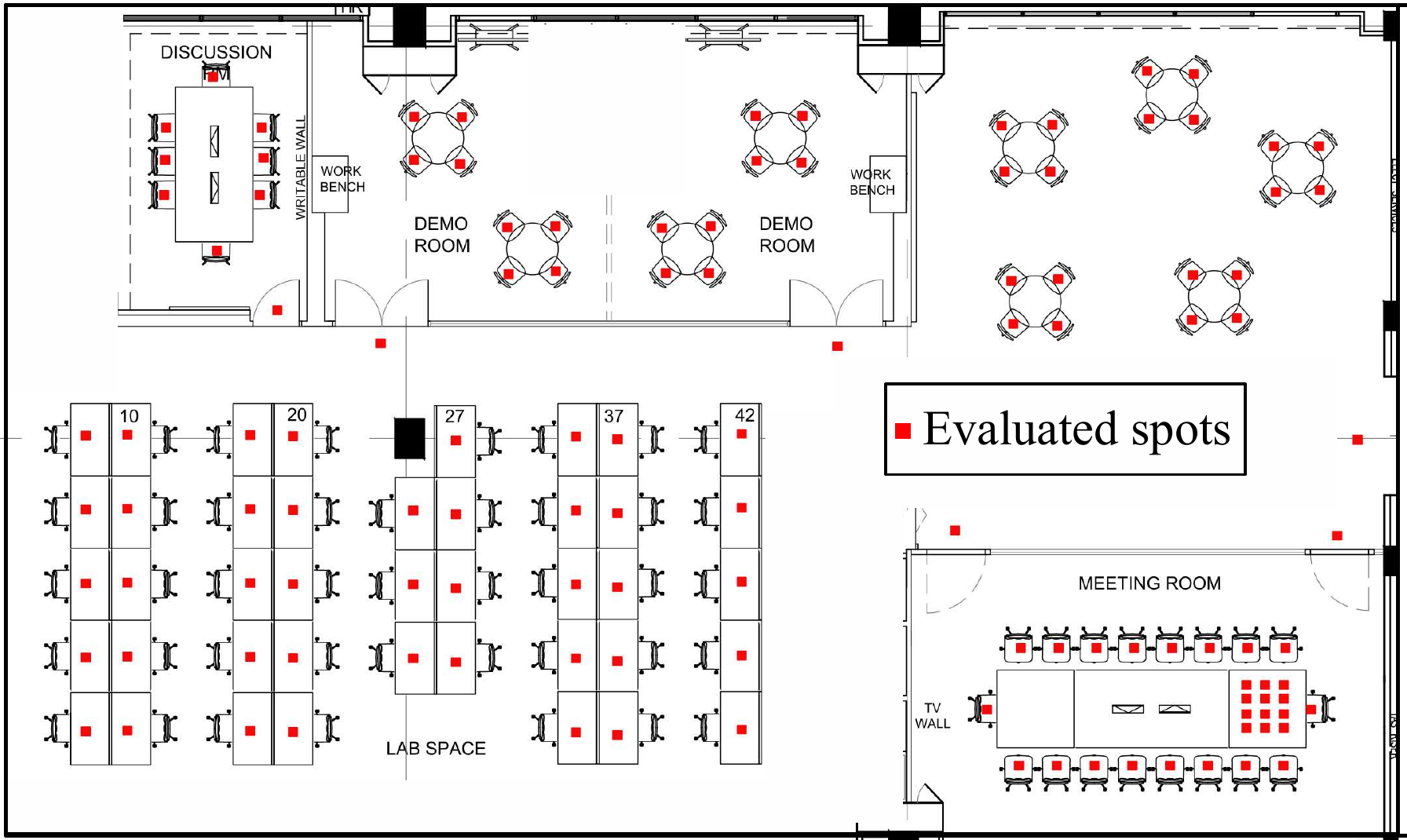}
  \vspace{-1em}
  \caption{Floor plan of the lab space.}
  \vspace{-1em}
  \label{fig:ncs-layout}
\end{figure}

$\blacksquare$ {\bf Echo extraction:} We develop a program to use the smartphone's loudspeaker to emit the excitation signal and microphone to record the echo at $44.1\,\text{ksps}$ for $100\,\text{ms}$. We assume the smartphone is held around 30 to $40\,\text{cm}$ in front of the chest. Note that the data collection cannot be unobtrusive, e.g., put the smartphone in a pocket. In the received data, we discard the first $10\,\text{ms}$ due to the propagation via the direct path from the loudspeaker to the microphone. 
We also discard the subsequent $1\,\text{ms}$ data, which usually contains the first reflection from the human body that is around $30 - 40\,\text{cm}$ away from the microphone. The subsequent $50\,\text{ms}$ data, which are collectively referred to as {\em echo trace} and illustrated in Fig.~\ref{fig:signal}b, are used for SLAM. Fig.~\ref{fig:signal}c shows the Pearson correlation between the received signal shown in Fig.~\ref{fig:signal}b and the chirp template. The peaks in Fig.~\ref{fig:signal}c indicate echoes.




\section{Measurement Study}
\label{sec:measure}

To gain insights for the system design, we conduct a set of measurement experiments in a $16 \times 28\,\text{m}^2$ lab space as shown in Fig.~\ref{fig:ncs-layout}.
We use a Google Pixel 4 smartphone to excite the space at 128 spots indicated by red squares in Fig.~\ref{fig:ncs-layout} and collect 1,700 echo traces at each spot. This section presents the analysis results for these traces.

\subsection{Spatial Distinctness of Echoes}



\begin{figure}
  \centering
  \begin{subfigure}{0.34\columnwidth}
    \centering
    \includegraphics[width=\columnwidth]{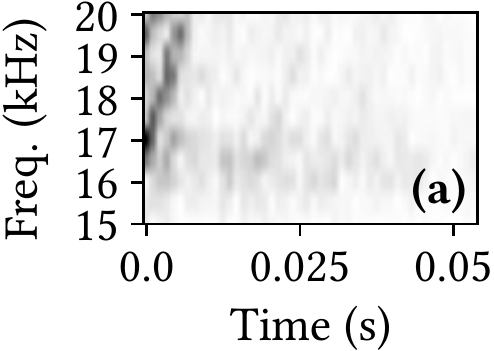} %
    \label{fig:near_far_a}
  \end{subfigure}
  \hfill
  \begin{subfigure}{0.28\columnwidth}
    \centering
    \includegraphics[width=\columnwidth]{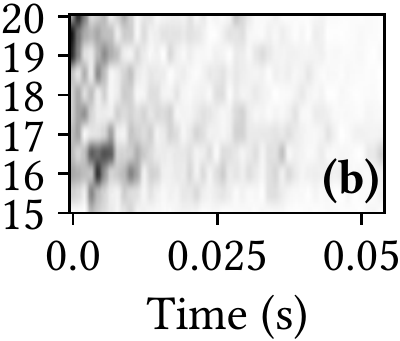} %
    \label{fig:near_far_b}
  \end{subfigure}
  \hfill  
  \begin{subfigure}{0.28\columnwidth}
    \centering
    \includegraphics[width=\columnwidth]{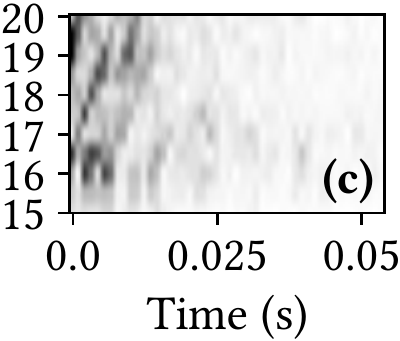} %
    \label{fig:near_far_c}
  \end{subfigure}  
  \vspace{-2em}
  \caption{Echo spectrograms obtained at three locations.}
  \label{fig:specs}
\end{figure}

\begin{figure}
  \centering
  \begin{subfigure}{0.49\columnwidth}
    \centering
    \includegraphics[width=0.78\columnwidth]{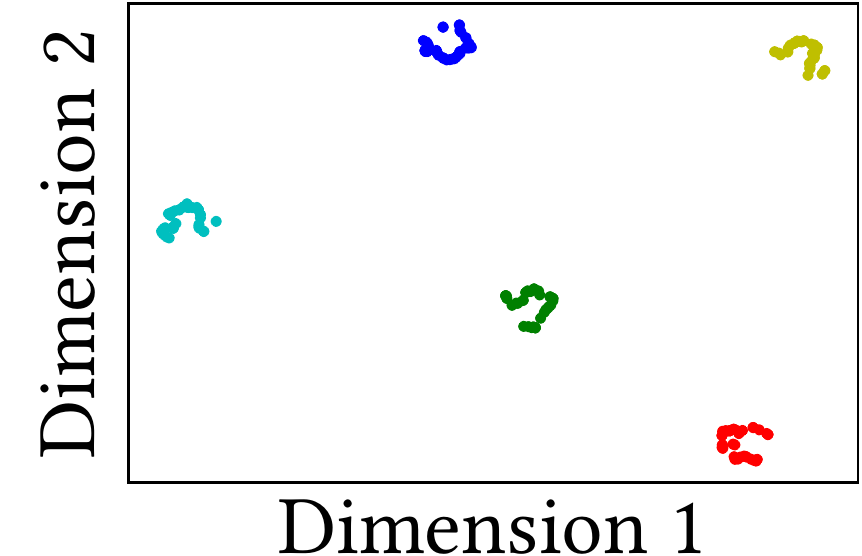}
    \caption{Chirp excitation}
    \label{fig:tsne-log}
  \end{subfigure}
  \begin{subfigure}{0.49\columnwidth}
    \centering
    \includegraphics[width=0.78\columnwidth]{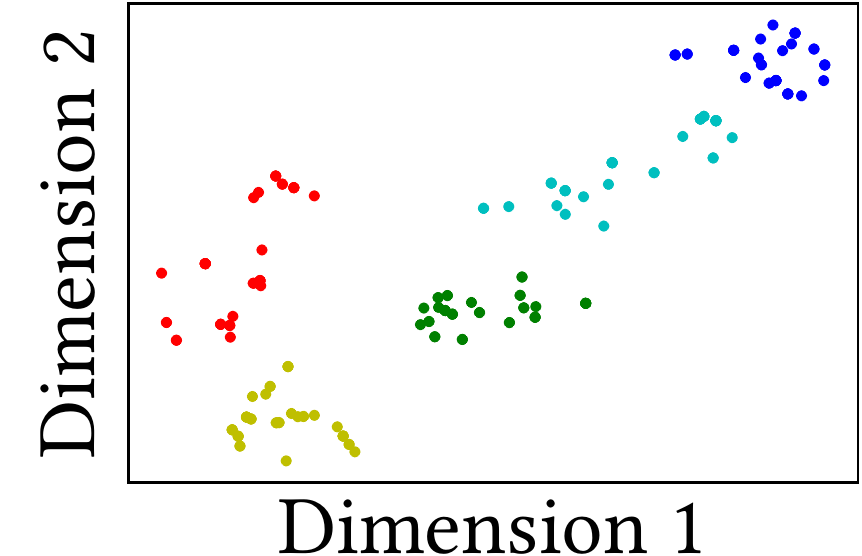}
    \caption{Single-tone excitation}
    \label{fig:tsne-20khz}
  \end{subfigure}
  \vspace{-1em}
  \caption{The t-SNE features of the echo spectrograms at five locations. Color represents locations.}
  \label{fig:tsne-result}
\end{figure}

We apply short-time Fourier transform (STFT) on the echo data to extract the spectrogram. Specifically, we slide a 96-point Hann window with 48 points of overlap on echo data, resulting in a $49 \times 48$ spectrogram. We further discard the frequency bins below $15\,\text{kHz}$, yielding a $12 \times 48$ image as the final result. Fig.~\ref{fig:specs} shows the echo spectrograms collected at three different spots.
The frequency dimension of a spectrogram is from $15$ to $20\,\text{kHz}$. The differences among the spectrograms suggest that echo traces vary across locations.
We also apply the t-distributed stochastic neighbor embedding (t-SNE) \cite{van2008visualizing} to reduce the dimension of the spectrograms collected at five spots with $1\,\text{m}$ separation in a linear topology.
Fig.~\ref{fig:tsne-result} shows the results when chirp and single-tone excitation signals are used. With the t-SNE features, we can visualize the spatial distinctness. From the figure, compared with single-tone excitation, the chirp excitation leads to more individually-compact and mutually-separated t-SNE feature clusters. This suggests that the chirp excitation brings more spatial distinctness of echo.


To understand the distinctness limit, we use the task of supervised learning-based localization to investigate the achievable spatial resolution and scalability with respect to the number of spots.



\subsubsection{Spatial resolution.}
\label{sec:spatial-resolution}
\begin{figure}
  \centering
  \begin{subfigure}{0.49\columnwidth}
    \centering
    \includegraphics[width=\columnwidth]{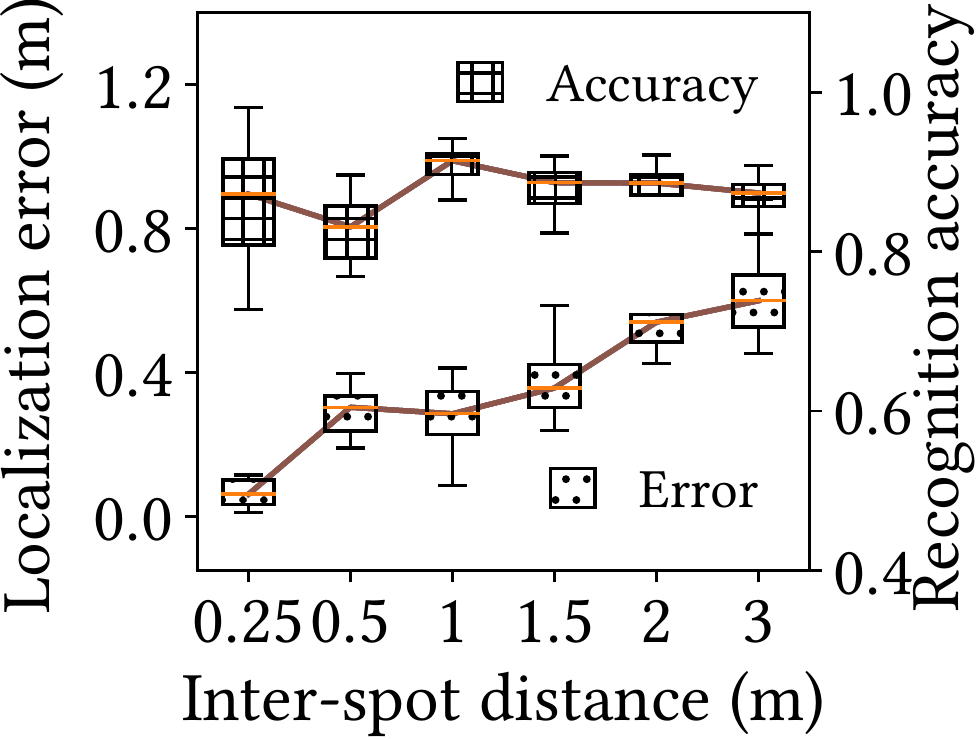}
    \vspace{-0.3em}
    \caption{Resolution}
    \label{fig:loc-resolution}
  \end{subfigure}
  \begin{subfigure}{0.49\columnwidth}
    \centering
    \includegraphics[width=\columnwidth]{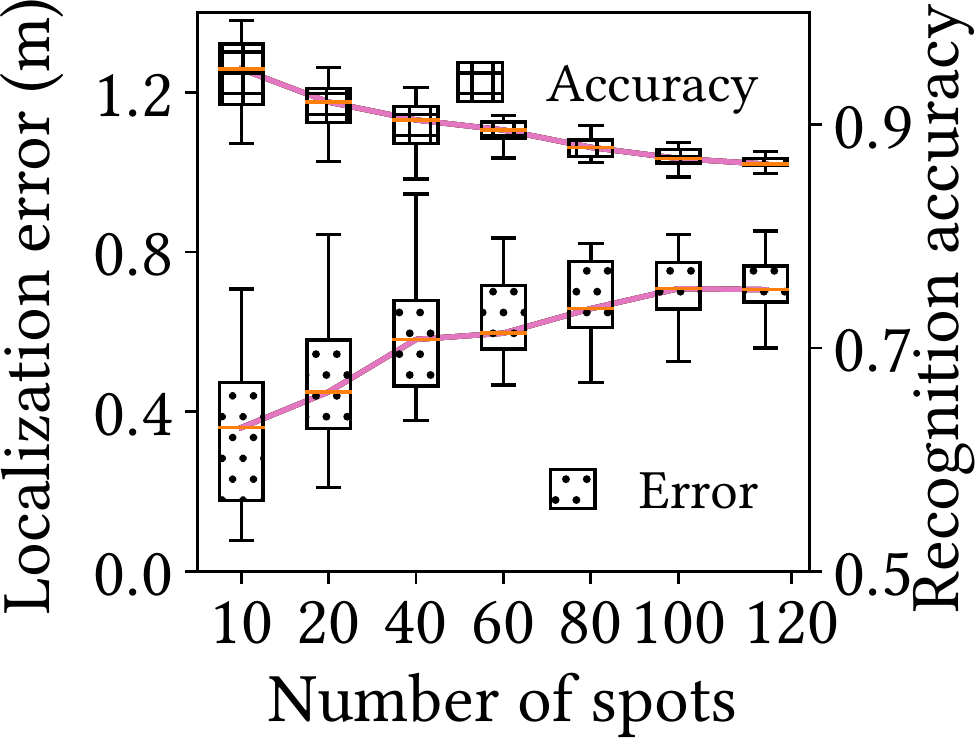}
    \vspace{-0.3em}
    \caption{Scalability}
    \label{fig:loc-scalable}
  \end{subfigure}
  \vspace{-1em}
  \caption{Distinctness limit of acoustic echoes.}
  \vspace{-1em}
\end{figure}

For each location, the spectrograms of the 1,700 echo traces are divided into 1,500 training samples and 200 test samples.
The ground truth labels correspond to the spot's location.
To understand how the inter-spot distance affects the localization accuracy, we divide the 128 spots into multiple groups with different densities. As a result, the average inter-spot distances of the groups range from $0.25\,\text{m}$ to $3\,\text{m}$. For each group, we use echo spectrograms to train a ResNet-18 DNN to classify the spots. We measure both the spot recognition accuracy and the mean localization error. Fig.~\ref{fig:loc-resolution} shows the measured results versus the average inter-spot distance. For each group, we repeat the evaluation process 20 times and plot the error bars.
The recognition accuracy remains at around 90\%.
The mean localization error increases with the inter-spot distance and remains at the sub-meter level.
The results suggest that the acoustic echoes can achieve sub-meter spatial resolution. 


\subsubsection{Scalability.} We increase the number of spots handled by a single ResNet-18 model (denoted by $k$) to understand the scalability. 
For each setting of $k$, we randomly draw $k$ spots from the 128 spots, train and test a ResNet-18 model. Note that the selected spots become denser for a larger $k$. We repeat the process 20 times for each $k$ setting.
Fig.~\ref{fig:loc-scalable} shows the results. The recognition accuracy gradually decreases with $k$ and becomes flat when $k$ exceeds 100.
This result is consistent with the intuition that the complexity of learning using a DNN increases with the number of classes.
The mean localization error increases with $k$ but remains under $1\,\text{m}$ and thus at the level of the spatial resolution limit shown in \sect\ref{sec:spatial-resolution}. 
The above results suggest that the ResNet-18 model does not present a bottleneck when the number of spots is up to 128. 


\subsection{Robustness of Acoustic Echoes}

\begin{figure}
  \subcaptionbox{Altitude\label{fig:measure-h}}{\includegraphics[width=0.35\columnwidth]{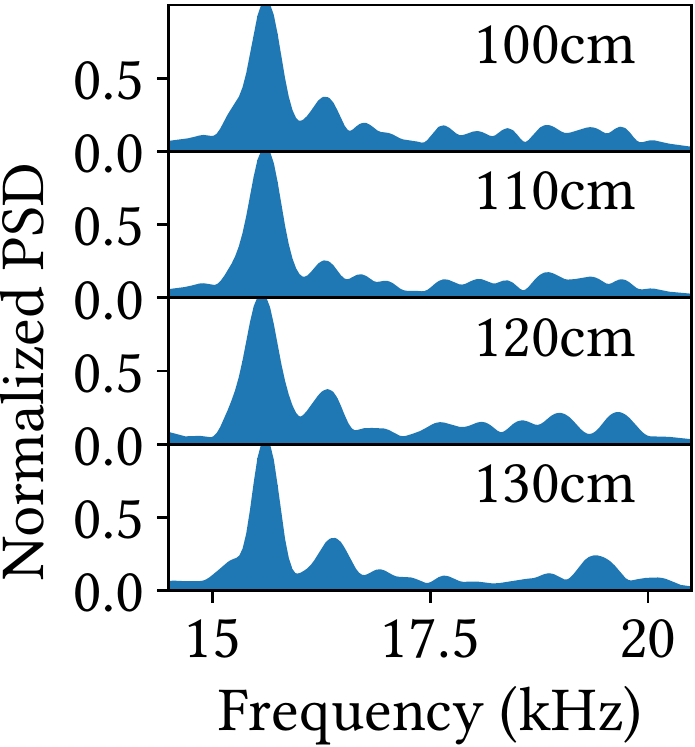}}\hfill
  \subcaptionbox{Orientation\label{fig:measure-o}}{\includegraphics[width=0.36\columnwidth]{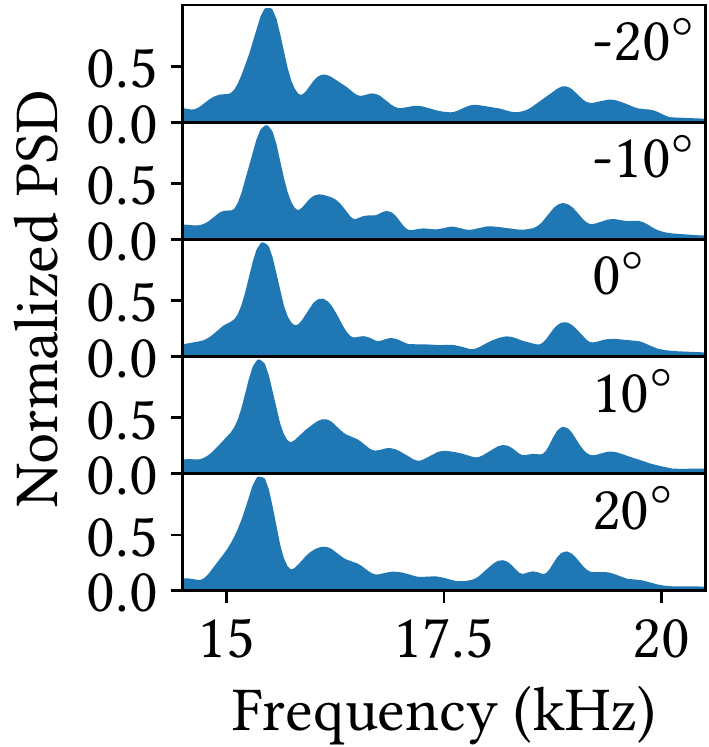}}\hfill
  \subcaptionbox{Data age\label{fig:measure-t}}{\includegraphics[width=0.27\columnwidth]{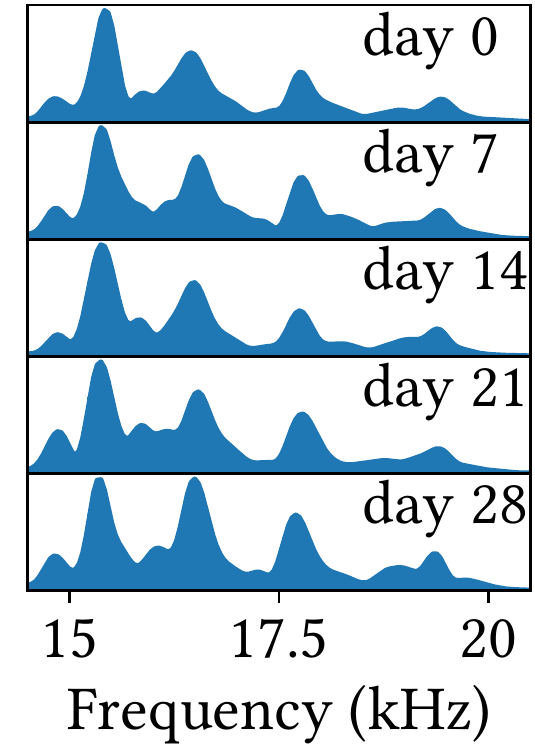}}\hfill
  \vspace{-1em}
  \caption{Power spectral densities (PSDs) of the acoustic echoes when several factors vary.}
  \label{fig:factors}
  \vspace{-1em}
\end{figure}
      
\begin{figure*}[ht]
  \includegraphics[width=2\columnwidth]{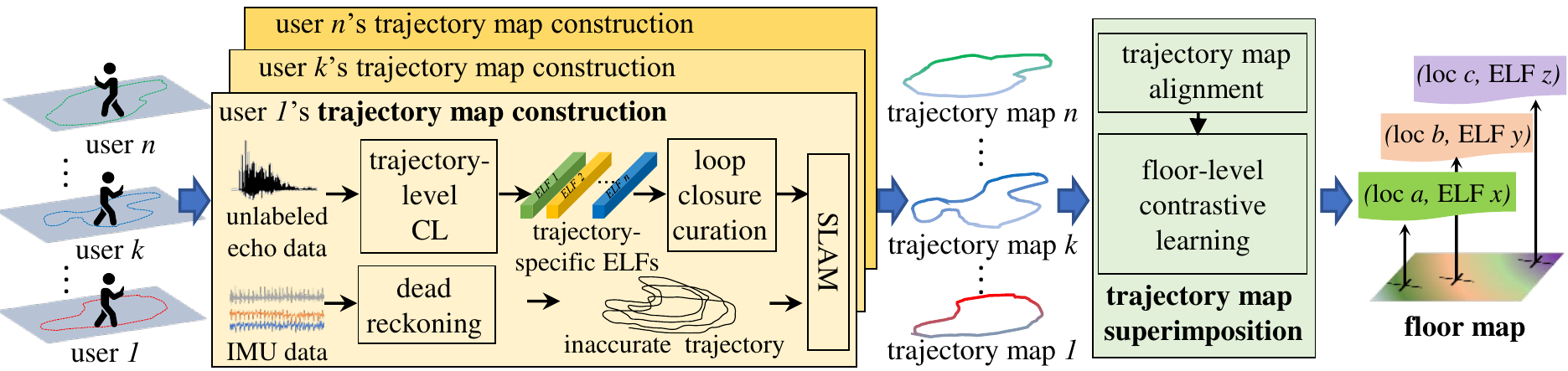}
  \vspace{-1em}
  \caption{Overview of the mapping phase of ELF-SLAM.}
  \label{fig:echoloc_workflow}
\end{figure*}


This section investigates the impacts of several potential affecting factors on acoustic echoes.

\subsubsection{Altitude.}
A researcher holds the phone at different altitudes to simulate the cases where users of different heights hold the phone with natural arm gestures for indoor navigation. 
As typical adult heights are within $150$--$194\,\text{cm}$~\cite{humanheight}, we test at the altitudes from $100\,\text{cm}$ to $130\,\text{cm}$ (i.e., two thirds of user height).
Fig.~\ref{fig:measure-h} shows the power spectral densities (PSDs) of the acoustic echoes at different altitudes of the same spot. We can see that the altitude variations of less than $30\,\text{cm}$ introduce little impact. Hence, the acoustic echoes are insensitive to variations of user height and hand altitude. 

\subsubsection{Phone orientation.}
As a smartphone's loudspeaker and microphone are not perfectly omnidirectional, phone orientation may affect the received signal at a spot.
We evaluate the stability of the received echoes when phone orientation varies. Fig. ~\ref{fig:measure-o} shows the echo PSDs when the phone has an orientation deviation of $-20^\circ$ to $20^\circ$ from a certain direction.
The results show that the echoes change slightly with less than $40^\circ$ orientation deviation.
When the orientation deviation is larger, the acoustic echoes exhibit larger differences.
This suggests that the impact of phone orientation needs to be properly dealt with when there are multiple echo traces collected at the same spot but in different phone orientations. We will present the solution to this issue in \sect\ref{sec:superimpose}.

\subsubsection{Temporal stability.}
We collect acoustic echoes at a certain location over a one-month period. In this period, the indoor layout has no significant changes. From Fig.~\ref{fig:measure-t}, the echo PSDs are consistent over the time period. In practice, the constructed floor map can be updated whenever a user contributes a trajectory map. The continuous update helps address the potential aging issue over longer periods of time. In \sect\ref{sec:measure}, we will evaluate the impact of significant changes of the indoor space layout (e.g., furniture re-arrangement) on our system and a mitigation approach beyond map update.





\section{Design of ELF-SLAM}
\label{sec:design}

From the measurement study, the acoustic echoes exhibit sub-meter spatial distinctness, which is the basis of the fingerprint approach. To unleash the fingerprint approach from laborious labeled training data collection, this section presents the design of ELF-SLAM based on acoustic echoes and IMU data captured by a smartphone during movements. In this section, 
\sect\ref{sec:overview} overviews the design of ELF-SLAM.
\sect\ref{sec:graph-slam} presents the graph-based SLAM formulation.
\sect\ref{sec:loop-closure} introduces ELF for loop closure detection.
\sect\ref{sec:loop-curation} presents a clustering-based approach for loop closure curation.
\sect\ref{sec:superimpose} presents the trajectory map superimposition.
\sect\ref{sec:loc-infer} presents ELF-based localization.

\subsection{Approach Overview}
\label{sec:overview}

As illustrated in Fig.~\ref{fig:echoloc_workflow}, the mapping phase of ELF-SLAM consists of two major components: {\em trajectory map construction} and {\em trajectory map superimposition}, where the former focuses on a single trajectory and the latter combines all available trajectories. The trajectories only cover a portion of the indoor space. However, when sufficient trajectories are collected, the combined map can cover the popular locations in the indoor space. 

$\blacksquare$ {\bf Trajectory map construction:} A user holds a smartphone and moves within the target indoor space to collect the acoustic echoes and IMU data simultaneously on the movement trajectory. The IMU data is used to reconstruct the user's trajectory via dead reckoning, while the acoustic echo data is used to detect loop closures on the user's trajectory. The detected loop closures provide critical regulation for the dead reckoning to deal with its long-run drifting problem. Since the generic acoustic features, such as PSD, spectrogram, etc, are ineffective for loop closure detection, ELF-SLAM applies CL to learn a custom and trajectory-specific ELF for loop closure detection. This trajectory-level CL consists of model {\em pre-training} using synthetic data from a room acoustics simulator and {\em fine-tuning} using real data collected by the user from the target indoor space. ELF-SLAM detects loop closures based on a custom similarity metric called echo sequence similarity (ESS) between two sequences of ELF traces. Then, a clustering-based approach removes the false positive detection results to curate the loop closures. Lastly, a graph-based SLAM algorithm constructs an accurate trajectory map of ELFs for the user.

$\blacksquare$ {\bf Trajectory map superimposition:} When multiple trajectory maps are available (e.g., through crowdsensing), they are superimposed to generate a floor map. The superimposition reconciles different trajectory maps' ELFs that are collected at the same spot but in different phone orientations. To achieve this, different users' trajectory maps are first aligned into a common coordinate system. The alignment can be achieved based on known initial positions of the users' trajectories (e.g., the entrance of the space) and/or prior knowledge about the accessible passages of the target indoor space \cite{montemerlo2002fastslam}. Then, we apply the floor-level CL to train a floor-wide ELF extractor using the acoustic data from all the trajectory maps. The floor map consists of the floor-level ELFs at all spots covered by all the trajectory maps, where each spot is associated with a single floor-level ELF. 

Once a map is available, a smartphone can be localized based on its captured echoes in response to a few chirps emitted by the phone. Specifically, the smartphone extracts the ELFs of the echoes and compares them against the map to estimate its location.

\subsection{Graph-based SLAM Formulation} 
\label{sec:graph-slam}
Graph-based SLAM \cite{grisetti2010tutorial} constructs a graph whose nodes represent the mobile's poses and edges represent the kinetic constraints relating two poses. When using IMU data to establish the kinetic constraints, loop closure is a vital regulation in combating the long-run drifting problem of IMU-based odometry \cite{randell2003personal}. In this paper, by letting $\mathbf{x}_{k}$ denote the node (i.e., location) corresponding to the $k^\text{th}$ detected footstep, the acoustic echo trace captured between the $k^\text{th}$ and $(k+1)^\text{th}$ footsteps is the measurement associated with the node $\mathbf{x}_{k}$ and used to detect whether $\vec{x}_k$ is at the same location as any previous node (i.e., loop closure detection). The edge connecting two nodes is associated with the IMU-based odometry. 
The user trajectory is estimated via the graph-based optimization after the loop closures are identified. The estimation method is as follow. For a total of $N$ detected footsteps, let $\mathbf{X}=\left\{\mathbf{x}_{1}, \ldots, \mathbf{x}_{N}\right\}$ denote the sequence of nodes that describes the user trajectory and $\mathbf{u}_{i,j}$ denote the edge constraint between nodes $\mathbf{x}_{i}$ and $\mathbf{x}_{j}$ based on the IMU data. 
Let $\mathcal{C}$ denote the set of footstep index pairs of the detected loop closures. The essence of the trajectory reconstruction can be described by the following optimization problem:
\begin{equation*}
  \mathbf{X}^{*}=\underset{\mathbf{X}}{\operatorname{argmin}}\!\!\!\!\!\!\! \sum_{\forall i \in [1, \ldots, N-1]} \!\!\!\!\!\!\!\!\! \left\|f\left(\mathbf{x}_{i}, \mathbf{u}_{i,i+1}\right)-\mathbf{x}_{i+1}\right\|^{2}+ \!\!\!\!\!\! \sum_{\forall \langle i, j \rangle \in \mathcal{C}} \!\!\!\!\!\! \left\|f\left(\mathbf{x}_{i}, \mathbf{u}_{i,j}\right)-\mathbf{x}_{j}\right\|^{2},
\end{equation*}

\noindent
where $\| \vec{x} - \vec{y}\|$ denotes the Euclidean distance between two locations $\vec{x}$ and $\vec{y}$, $f(\vec{x}_i, \vec{u}_{i,j})$ represents the prediction of $\vec{x}_j$ based on $\vec{x}_i$ and $\vec{u}_{i,j}$. 
In this paper, we implement the SLAM algorithm based on the general graph optimization framework in \cite{kummerle2011g}, which also addresses the uncertainty of the prediction $f(\vec{x}_i, \vec{u}_{i,j})$. 
  
\subsection{ELF For Loop Closure Detection}
\label{sec:loop-closure}

\begin{figure}[t]
  \includegraphics[width=\columnwidth]{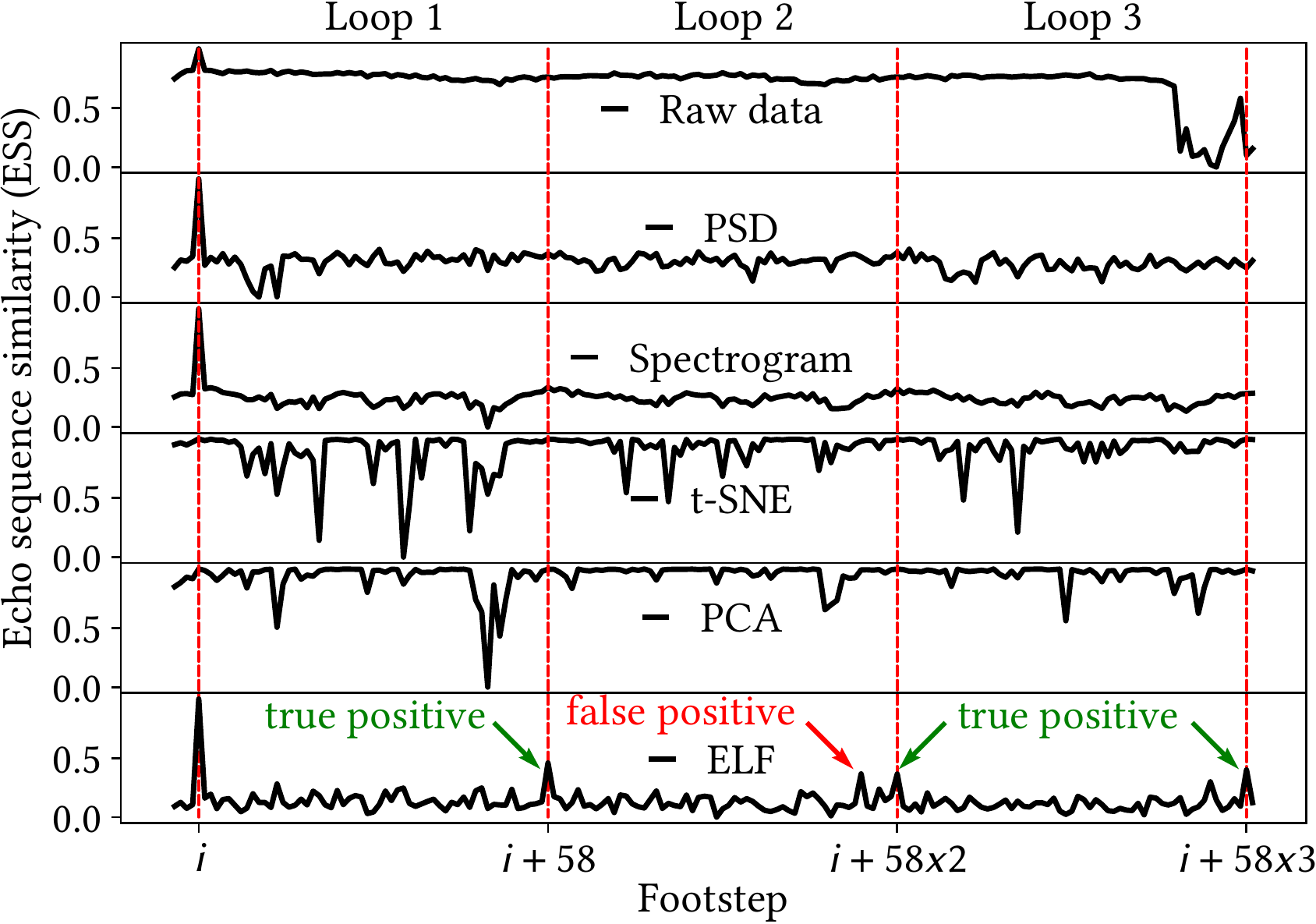}
  \vspace{-2em}
  \caption{ESS traces with respect to footstep $i$. Peaks indicate loop closures at footstep $i+58$, $i+58 \times 2$, and $i+58 \times 3$.}
  \label{fig:feat-d}
  \vspace{-1em}
\end{figure}

Identifying an effective feature for loop closure detection is critical to SLAM. In this section, we first demonstrate the ineffectiveness of the generic features. Then, we propose using CL to construct a learning-based feature.

\subsubsection{Ineffectiveness of generic features.} We conduct a controlled experiment to evaluate several generic acoustic features. 
A researcher walks four rounds by following 58 markers pasted on the floor of the lab shown in Fig.~\ref{fig:ncs-layout} and uses a phone to collect acoustic data. As such, each round consists of exact same 58 footsteps. We compute the following features of the echo data: PSD, spectrogram, t-SNE, and principal component analysis (PCA). Then, we compute the similarity between the features collected at footstep $i$ in the first round with those at all footsteps in the four rounds. The similarity is measured by the echo sequence similarity (ESS), which is defined as follows. 
For two footsteps $i$ and $j$ at which $K_i$ and $K_j$ features are collected, the ESS between them is obtained by averaging the $K_{i} \times K_{j}$ pair-wise cosine similarity among the two sets of features. 
Fig.~\ref{fig:feat-d} shows the resulting ESS traces with footstep $i$ (where $i = 4$) in the first round and $j$ being all footsteps of the four rounds sequentially. In this controlled experiment, with respect to the footstep $i$, loop closures are formed at the footsteps $i+58$, $i+58 \times 2$, and $i+58 \times 3$. If the used feature is effective, ESS peaks should be observed at these footsteps. However, from the plots in the first five rows of Fig.~\ref{fig:feat-d}, no salient peaks are observed at these footsteps. 
This suggests that the raw data and the used generic feature extraction techniques are ineffective for our loop closure detection problem. Note that although t-SNE is effective for finding feature embeddings of clustered data \cite{arora2018analysis} as shown in Fig.~\ref{fig:tsne-log}, it is ineffective on the echo samples collected in the spatial continuum that do not exhibit clustered patterns. Note that the ESS peaks are also not observed under other similarity metrics, e.g., those related to Euclidean and Manhattan distances, etc. 

\subsubsection{Learning-based ELF.}
\label{sec:learn-elf}

\begin{figure}
  \centering
  \includegraphics[width=\columnwidth]{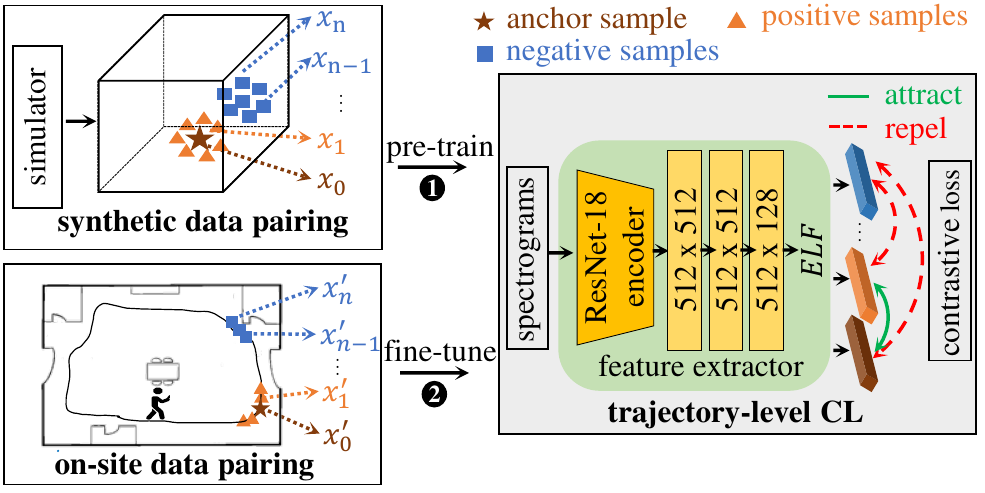}
  \caption{Trajectory-level CL to learn trajectory-specific ELFs.}
  \label{fig:loop-cl}
\end{figure}

The ineffectiveness of generic features motivates us to apply CL to construct a custom feature, i.e., ELF. 
In what follows, we present our CL design to construct the ELF for loop closure detection. Fig.~\ref{fig:loop-cl} depicts the workflow. It consists of three steps: {\em data pairing}, {\em model pre-training}, and {\em model fine-tuning}.

\begin{figure}
  \centering
  \begin{subfigure}{0.33\columnwidth}
    \centering
    \includegraphics[width=\columnwidth]{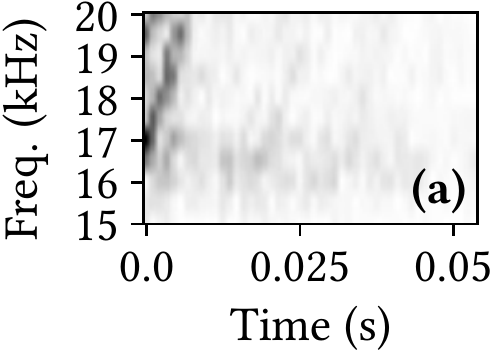} %
    \label{fig:near_far_a}
  \end{subfigure}
  \hfill
  \begin{subfigure}{0.28\columnwidth}
    \centering
    \includegraphics[width=\columnwidth]{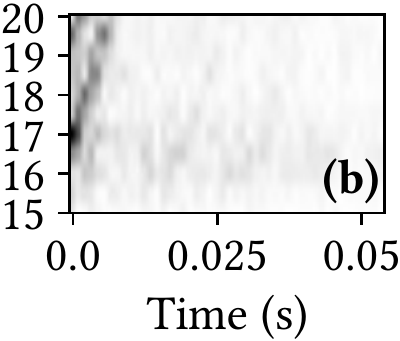} %
    \label{fig:near_far_b}
  \end{subfigure}
  \hfill  
  \begin{subfigure}{0.28\columnwidth}
    \centering
    \includegraphics[width=\columnwidth]{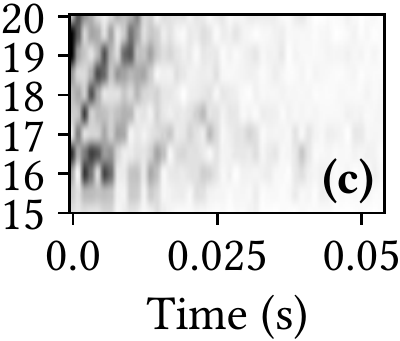} %
    \label{fig:near_far_c}
  \end{subfigure}  
  \vspace{-1em}
  \caption{Spectrograms of echoes received at three spots. The spots of (a) and (b) are $10\,\text{cm}$ apart from each other; the spot of (c) is $2\,\text{m}$ apart from those of (a) and (b).}
  \vspace{-1em}
  \label{fig:spec-nearfar}
\end{figure}

{\bf Data pairing} constructs positive/negative data pairs needed by CL. In image recognition tasks, the positive samples are constructed by introducing spatial perturbations such as resizing, cropping, and blurring, which do not erase the information needed for image recognition. 
However, such spatial perturbations are not applicable to the echo data, because they destruct the subtle structural information embedded in the echo signal that is related to the smartphone's location. 
The design of our data pairing approach is based on the observation that the echoes are similar if collected at close locations and distinct if collected at locations apart. This is illustrated using real data shown in Fig.~\ref{fig:spec-nearfar}. 
Thus, we construct positive pairs of echoes collected at close locations and negative pairs of echoes at locations apart. Specifically, we treat two consecutive echoes as a positive pair. 
For each training step, we randomly sample a training batch of 256 such positive pairs from the entire sequence of echoes collected by a certain user. According to our design in \sect\ref{sec:measure}, the time gap between two consecutive echoes is $0.1\,\text{s}$. As the average human walking speed is $5\,\text{km/h}$, the locations for collecting two consecutive echoes are separated by $0.14\,\text{m}$ on average. This average separation is slightly lower than the achievable spatial resolution of the echo modality as evaluated in \sect\ref{sec:measure}. Thus, viewing two consecutive echoes during the user's movement as a positive pair is a good heuristic. Then, the cross pairing of the members of the 256 positive pairs gives the negative pairs. Note that the chance of wrongly forming a negative pair with two echo samples collected at a loop closure is low. Such limited wrong forming of negative pairs will not devastate the CL. 

{\bf Model pre-training} exploits self-supervised learning to build a basic ELF extractor, which will be specialized by the model fine-tuning step presented later. 
Self-supervised learning often requires abundant unlabeled training data for feature representation learning. To reduce the overhead of data collection in real environments, we use \texttt{pyroomacoustic} \cite{scheibler2018pyroomacoustics}, which is a room acoustics simulator, to generate abundant synthetic training data. Specifically, we simulate a smartphone with a speaker and a microphone separated by $15\,\text{cm}$ and generate massive echo data at fine-grained grid points in various simulated rooms with different shapes and sizes. The synthetic echo samples are used to train a feature extractor DNN by CL. We adopt a feature extractor architecture from \cite{chen2020big}, which consists of a ResNet-18 encoder and a 3-layer projecction head. The extractor maps the input spectrogram of an echo trace to a 128-dimensional ELF. As the locations of the synthetic echoes are controlled and thus known, we treat the synthetic echoes collected at locations separated by less than $20\,\text{cm}$ as positive pairs, and the rest as negative pairs. The following contrastive loss adopted from \cite{chen2020simple} is minimized during the model pre-training:
\begin{equation*}
  \ell_{i, j}=-\log \frac{\exp \left(\operatorname{sim}\left(\boldsymbol{z}_{i}, \boldsymbol{z}_{j}\right) / \tau\right)}{\sum_{k=1}^{2 M} \mathbf{1}_{[k \neq i]} \exp \left(\operatorname{sim}\left(\boldsymbol{z}_{i}, \boldsymbol{z}_{k}\right) / \tau\right)},
\end{equation*}
\noindent
where $\mathbf{1}_{[k \neq i]} \in\{0,1\}$ is evaluated to 1 if and only if $k \neq i$, $\mathrm{sim}(\cdot, \cdot)$ denotes cosine similarity, $\boldsymbol{z}$ is the extracted feature vector, $i$ and $j$ indicate a positive pair, $M$ is batch size, and $\tau$ is the temperature parameter. With the above contrastive loss, the pre-training increases the feature similarity for echoes at close locations and decreases the feature similarity for those at locations apart. Thus, it builds a feature extractor that discriminates the echoes' locations under the cosine similarity metric. As a result, the loop closure detection can be implemented by comparing the ELFs produced by the feature extractor in terms of the cosine similarity. The necessity of the model pre-training will be evaluated in \sect\ref{sec:eval}. 
 
{\bf Model fine-tuning} uses a small amount of unlabeled data collected by users in a specific target space to adapt the pre-trained model to capture the environment-specific characteristics. 
We follow the same self-supervised CL procedure as described above and construct the data pairs to fine-tune the feature extractor. The resulted feature extractor can generate trajectory-specific ELFs for loop closure detection in the target space.


\begin{figure}[t]
	\begin{subfigure}[c]{.48\columnwidth}
		\includegraphics[width=\columnwidth]{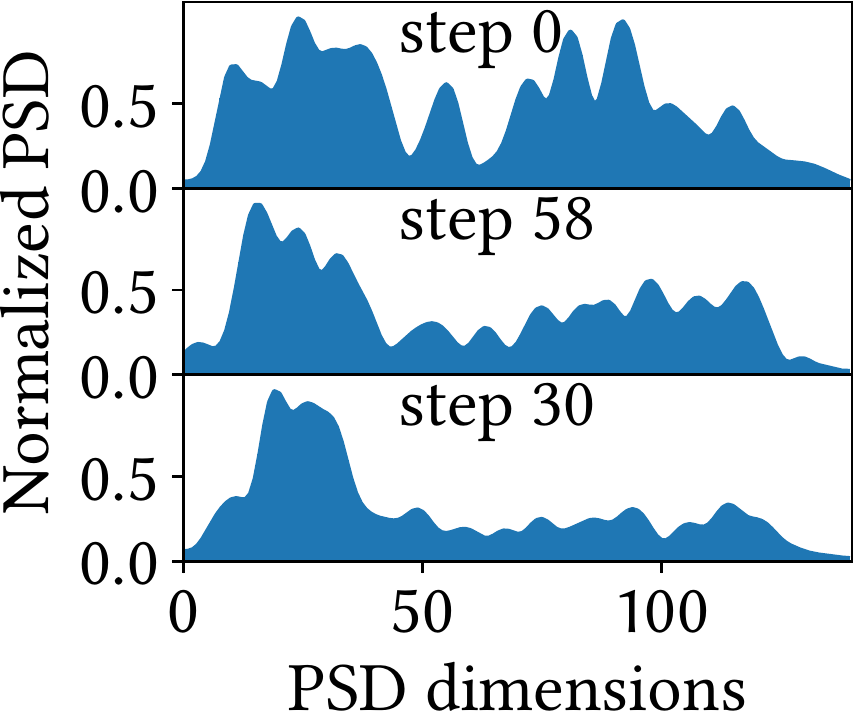}
		\caption{$\mathrm{sim}(0,58)=0.40, \mathrm{sim}(0,30)=0.42$}
		\label{fig:traj_psd}
		\end{subfigure}
  \hfill
  \hspace{0.5em}
	\begin{subfigure}[c]{.48\columnwidth}
	    \includegraphics[width=\columnwidth]{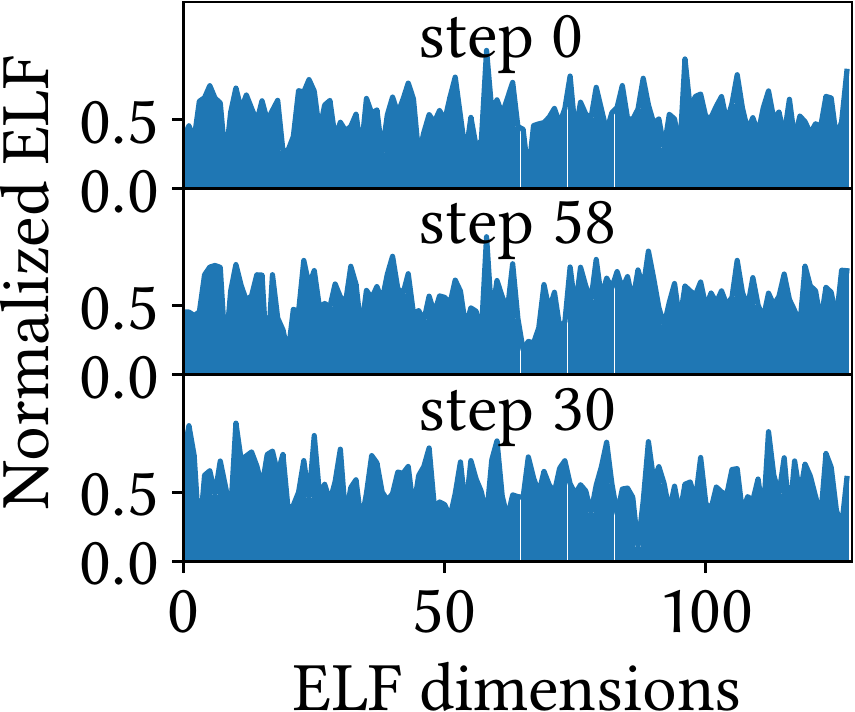} 
	    \caption{$\mathrm{sim}(0,58)=0.61, \mathrm{sim}(0,30)=0.22$}
	    \label{fig:traj_elf}
	\end{subfigure}
	\caption{(a) PSDs and (b) ELFs visualization at steps 0, 30, and 58.}
	\label{fig:traj-elf}
\end{figure}

\subsubsection{Loop closure detection using ELF.}

The last row of Fig.~\ref{fig:feat-d} shows the ESS trace computed using ELF. It shows peaks at footstep $i+58$, $i+58 \times 2$, and $i+58 \times 3$ as marked by the green arrows. Fig.~\ref{fig:traj-elf} shows the cosine similarities using echo features from steps 0, 30 and 58. For PSD, the cosine similarity between loop closure steps 0 and 58 is 0.4, and between the non-loop closure steps 0 and 30 is 0.42. PSD cannot differentiate the loop closure and non-loop closure data. Thus, PSD is ineffective for loop closure detection. For ELF, the cosine similarity between loop closure steps 0 and 58 is 0.61, and between the non-loop closure steps 0 and 30 is 0.22. The learned ELFs can see significant differences on the loop closure and non-loop closure data. Thus, ELF is effective for loop closure detection.

However, we also observe an unexpected peak close to the footstep $i+58 \times 2$ as marked by a red arrow, which may lead to a false positive loop closure detection in Fig.~\ref{fig:feat-d}. Unfortunately, the SLAM is often sensitive to false positive loop closures -- some false positives can degrade the SLAM performance \cite{lu2018simultaneous}. Thus, a loop closure curation algorithm is needed to remove the false positives. 

\subsection{Loop Closure Curation}
\label{sec:loop-curation}

\subsubsection{Approach design.}
We propose a clustering-based loop closure curation approach that is based on an ESS matrix defined as follows.

\begin{figure}
  \centering
  \begin{minipage}{.39\columnwidth}
    \centering
    \includegraphics[width=\columnwidth]{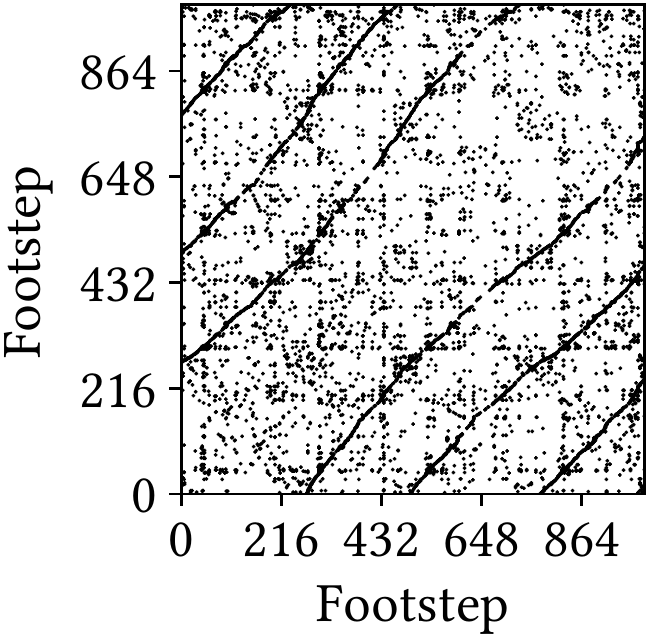}
    \caption{ESS matrix with true loop closures forming trend curves.}
    \label{fig:d_b}
  \end{minipage}
  \hspace{0.5em}
  \begin{minipage}{.49\columnwidth}
  \centering
  \includegraphics[width=0.96\columnwidth]{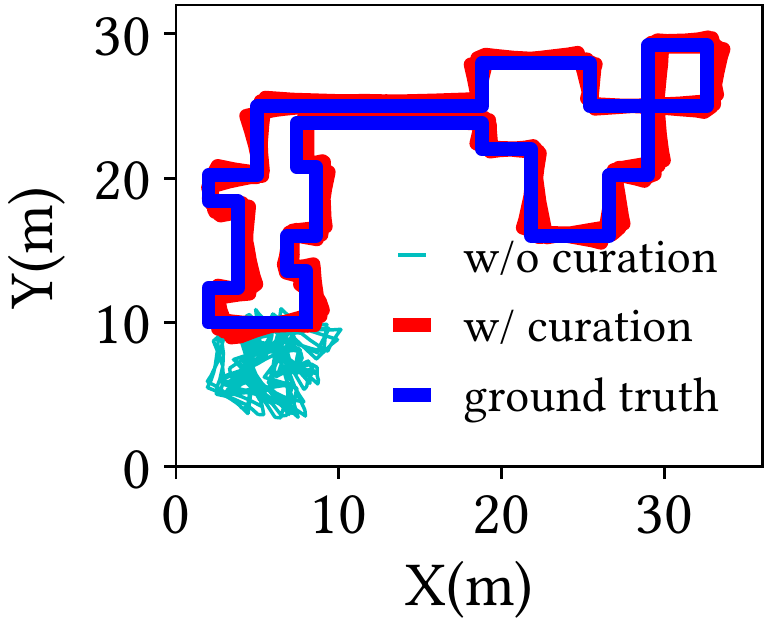}
  \caption{Reconstructed trajectories with/without loop closure curation.}
  \label{fig:d_a}    
  \end{minipage}
\end{figure}

{\bf ESS matrix: } Consider a user's trajectory consisting of $N$ footsteps. The pair-wise ESSs between any two footsteps form a $(N - 1) \, \times (N - 1)$ ESS matrix, where the $(i,j)^\text{th}$ element is the ELF-based ESS between the footsteps $i$ and $j$. Thus, the ESS matrix is symmetric. A large ESS suggests a high similarity between the two involved footsteps' ELFs, signaling a potential loop closure. Our extensive experiments show that a threshold value of $0.4$ for ESS can identify most true positive loop closures while capturing an acceptably low number of false positives to be removed shortly. The ESS matrix is binarized using the threshold, where the positive elements represent candidate loop closures. Fig.~\ref{fig:d_b} shows an example of the binarized ESS matrix constructed using the ELFs collected in a shopping mall. The x- and y-axis represent the footstep index. The black dots in the ESS matrix represent the positive elements. 

\begin{figure}
  \centering
  \includegraphics[width=\columnwidth]{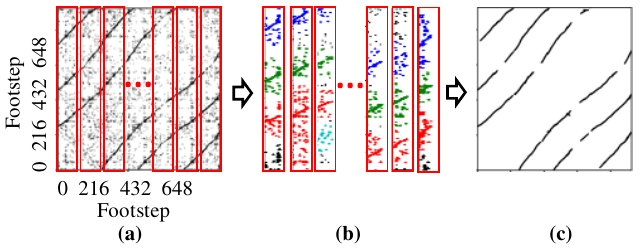}
  \caption{Clustering-based approach for loop closure curation: (a) ESS matrix slicing, (b) clustering in each slice, and (c) concatenated line regression results.}
  \label{fig:clustering}

\end{figure}

{\bf Clustering-based approach for loop closure curation: } 
The goal of loop closure curation is to remove the false positives from the binarized ESS matrix. Due to the user movement, the true positives in the binarized ESS matrix form trend curves. For instance, consider an ideal case in which the user walks at a constant speed, the true loop closures of footsteps 0, 1, ... , and 10 are footsteps $0 + L$, $1 + L$, ... , and $10 + L$, where $L$ is the loop length. As a result, the $\left( 0,0 + L \right)^\text{th}$, $\left( 1,1 + L \right)^\text{th}$, ... , and $\left( 10,10 + L \right)^\text{th}$ elements of the binarized ESS matrix should be positives and form a trend line. In contrast, the false positives tend to appear at random positions in the ESS matrix, as shown in Fig.~\ref{fig:d_b}. Based on this observation, we propose a clustering-based approach to isolate the trend curves formed by the true positives from the scattered false positives.

First, we divide the binarized ESS matrix into multiple slices, as illustrated in Fig.~\ref{fig:clustering}a. With the slicing, it is easier to identify the true positive clusters in each slice. In our implementation, we set the slice width to be 16 footsteps. Then, for the positives in each slice, we apply the DBSCAN clustering algorithm \cite{ester1996density} to identify the number of loops and divide the positives into multiple clusters. This is illustrated by Fig.~\ref{fig:clustering}b, where the clusters are differentiated by colors. Although some false positives are classified by DBSCAN as outliers, the remaining false positives close to the trend curves are still in the clusters. To remove these remaining false positives, for the points in each cluster, we apply the random sample consensus (RANSAC) \cite{fischler1981random} linear regression algorithm to detect a line approximating the trend curve segment. RANSAC is a preferred regression algorithm when many outliers are present. Concatenation of the regressed line segments across all slices gives the clean trend curves. Fig.~\ref{fig:clustering}c shows the concatenated results, in which the trend curves formed by the positives are effectively isolated from the scattered noises as shown in Fig.~\ref{fig:clustering}a. 
Lastly, we further curate the loop closure candidates based on the ESS matrix's symmetric property. Since the negative impact of a false positive on SLAM outweighs that of a false negative, we apply a strategy of only retaining the positives that conform to the symmetric property. Specifically, if the positive at the $(i,j)^\text{th}$ position of the ESS matrix has no counterpart positive at the $(j,i)^\text{th}$ position, the positive is viewed false and excluded.


\subsubsection{Effectiveness of loop closure curation.}

To demonstrate the impact of the false positives on SLAM, we use all positives in Fig.~\ref{fig:d_b} as the loop closure information to construct the trajectory map. The plot labeled "w/o curation" in Fig.~\ref{fig:d_a} shows the trajectory of the constructed map. We can see that the false positives devastate the trajectory reconstruction. The plot labeled "w/ curation" in Fig.~\ref{fig:d_a} shows the reconstructed trajectory using the curated loop closures. The new trajectory highly resembles the ground truth. The result demonstrates the effectiveness of the proposed clustering-based loop closure curation.

\subsection{Trajectory Map Superimposition}
\label{sec:superimpose}

\begin{figure*}
	\begin{subfigure}[c]{.38\columnwidth}
	  \includegraphics[width=\columnwidth]{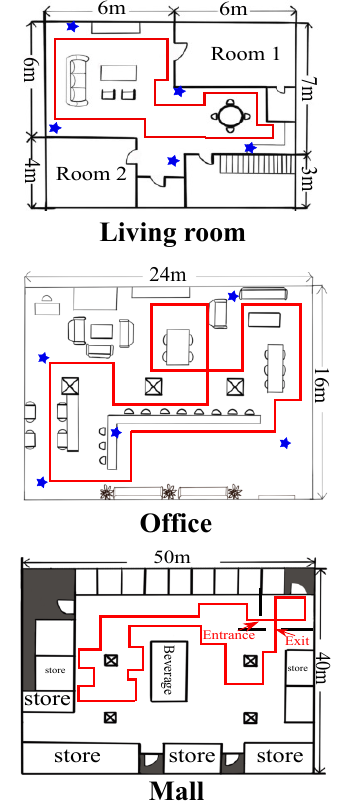}
	  \caption{Floor plans}
	  \label{fig:map-floor-plan}
	\end{subfigure}
	\hfill
	\begin{subfigure}[c]{.42\columnwidth}
	  \includegraphics[width=\columnwidth]{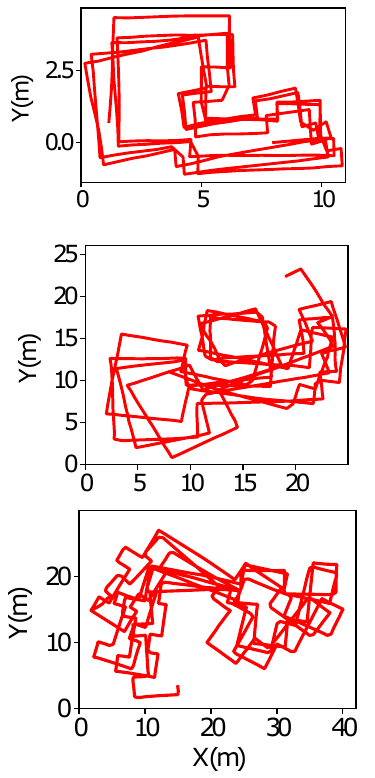}
	  \caption{IMU}
	  \label{fig:map-IMU}
	\end{subfigure}
  \hfill
	\begin{subfigure}[c]{.40\columnwidth}
	  \includegraphics[width=\columnwidth]{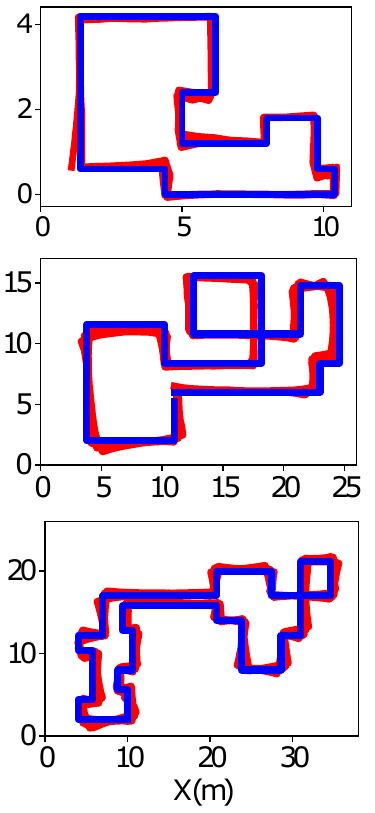}
	  \caption{ELF}
	  \label{fig:map-ELF}
	\end{subfigure}
	\hfill
	\begin{subfigure}[c]{.39\columnwidth}
	  \includegraphics[width=\columnwidth]{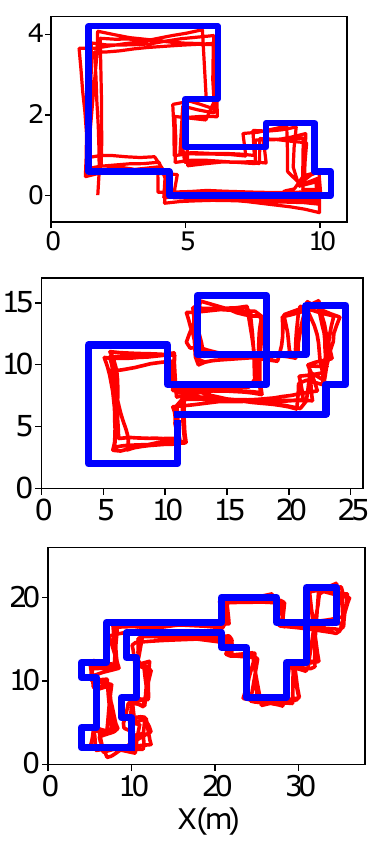}
	  \caption{Wi-Fi}
	  \label{fig:map-wifi}
	\end{subfigure}
	\hfill
	\begin{subfigure}[c]{.39\columnwidth}
	  \includegraphics[width=\columnwidth]{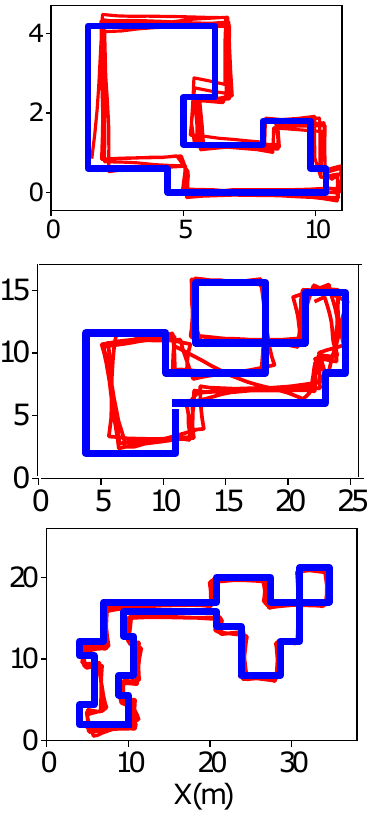}
	  \caption{Geomagnetism}
	  \label{fig:map-geo}
	\end{subfigure}
	\caption{Floor plans and trajectory reconstruction results in three evaluation environments.}
	\label{fig:map-results}
\end{figure*}

The trajectory map constructed from a single user's data only contains the echo data on a specific trajectory. For real applications, it is desirable to combine many trajectory maps to form a floor map that covers most/all accessible locations. We assume that each trajectory map's initial position relative to the indoor space is known. For instance, the SLAM mobile app may prompt the user to start the process from the entrance of the indoor space. When there are multiple entrances, location tagging \cite{tung2015echotag} can be used to recognize the actual entrance. With the known initial position, the trajectory maps can be collated onto a common coordinate system. However, the trajectory maps crossing a certain spot from different entering directions may have different echo data for the spot, due to the dependency of echo data on phone orientation as shown in \sect\ref{sec:measure}. Thus, we need to reconcile such differences. 

We propose a floor-level CL approach to train a unified feature extractor for map superimposition. It shares the same model pre-training workflow as the trajectory-level CL except the data pairing approach for model fine-tuning. Specifically, the echo data collected at the same location regardless of the phone orientation are treated as positive pairs, whereas those collected from different locations are treated as negative pairs. The feature extractor trained via the floor-level CL generates the floor-level ELFs covering all trajectory maps. As the quality of the floor map is related to its spatial coverage, this floor-level CL approach needs to scale well with the number of locations. In \sect\ref{sec:eval}, we evaluate this approach in handling 4,000 different locations with four phone orientations at each location. 

When falling back to the scheme of learning a location recognition model in the supervised manner, a possible approach to mitigate the echo data's sensitivity on phone orientation is to form a training dataset with echo data and location labels (regardless of orientation) from all the trajectory maps. In \sect\ref{sec:eval}, we will compare the localization performance of this supervised learning approach with our approach based on the floor map. 

\subsection{Localization}
\label{sec:loc-infer}

Once a map (either a trajectory map or floor map) is constructed, a smartphone's location can be determined after capturing the echoes in response to the chirps. We consider two localization approaches, i.e., \textit{one-shot localization} and \textit{trajectory localization}, which are suitable for the scenarios where the user stands still and moves, respectively. In the former, an ELF sequence containing multiple consecutive echoes collected at a spot is matched against the map in terms of the ESS to determine the location. In the latter, both the ELF sequence and the IMU data during the user's movement over a short time period are used for localization. Specifically, we apply dead reckoning to the IMU data to estimate the user's trajectory, and then apply a curve matching algorithm \cite{cui2009curve} to find the candidate segments in the map that resemble the user's trajectory. The candidate segment that has the largest average ESS from the captured ELF sequence is the output of the trajectory localization. 

\section{Performance Evaluation}
\label{sec:eval}

\subsection{Experiment Setup}

{\bf Evaluation environments: } we evaluate ELF-SLAM in three indoor environments: a living room ($60 \, \text{m}^{2}$), an office ($360 \, \text{m}^{2})$, and a shopping mall ($2,000 \, \text{m}^{2}$). The floor plans are shown in Fig.~\ref{fig:map-floor-plan}. To conduct comparative evaluation side by side, we employ the SLAM systems using two smartphone's built-in sensing modalities, i.e., Wi-Fi RSSI and geomagnetism, as the baselines. This is the same as the evaluation methodology adopted in \cite{lu2018simultaneous} that studies powerline EMR SLAM. Note that we also compare the results of ELF-SLAM and EMR SLAM. To implement Wi-Fi SLAM, we deploy five Wi-Fi access points (APs) in the living room and office, as illustrated by the stars in Fig.~\ref{fig:map-floor-plan}, that provide sufficient Wi-Fi coverage. The large shopping mall has dense APs deployed by the tenants. The number of Wi-Fi APs observable is around 5 to 10 when conducting experiments in the mall. Random people are walking in the shopping mall when we collect the data.

{\bf Data collection: } We develop an Android app and run it on a Google Pixel 4 smartphone to collect acoustic echoes, Wi-Fi RSSI, geomagnetic field signals, and IMU data. 
The app uses the \texttt{WifiManager} Android API to scan the surrounding Wi-Fi APs and collect RSSI data at a sampling rate of $0.8\,\text{sps}$. We do not sample Wi-Fi channel state information (CSI), because CSI sampling requires rooting a phone \cite{hernandez2020performing}. The app also samples the phone's built-in magnetometer at $50\,\text{sps}$. 
During data collection, the user holds the phone with one hand around 30 to $40\,\text{cm}$ in front of the chest and walks on a marked trajectory for multiple rounds in each of the evaluation environments.  
Note that the purpose of trajectory marking is to obtain the location ground truth.

{\bf Loop closure detection for baseline modalities: } 
For Wi-Fi SLAM, we use the Euclidean distance between two Wi-Fi RSSI vectors for loop closure detection \cite{yang2012locating}. For geomagnetic SLAM, we first normalize the triaxial magnetic data and then use dynamic time warping distance as the metric for loop closure detection \cite{wang2016keyframe}. We also apply the loop closure curation and graph-based optimization algorithms described in \sect\ref{sec:design} to the baseline SLAM systems.
\begin{table}
	\caption{Mapping error statistics.}
	\label{tab:map-stats}
	\begin{threeparttable}
	\footnotesize
	\centering
	\begin{tabular}{c|ccc|ccc|ccc}
	  \hline
	  \multirow{2}{*}{Modality} & \multicolumn{3}{c|}{Living room} & \multicolumn{3}{c|}{Office} & \multicolumn{3}{c}{Mall}\\
	   & $\tilde{x}$ \tnote{1} & $\bar{x}$ \tnote{2} & Q3 \tnote{3} & $\tilde{x}$ & $\bar{x}$ & Q3 & $\tilde{x}$ & $\bar{x}$ & Q3 \\
	  \hline
	  {\bf ELF} & {\bf 0.10} & {\bf 0.10} & {\bf 0.14} & {\bf 0.63} & {\bf 0.63} & {\bf 0.80} & {\bf 0.45} & {\bf 0.53} & {\bf 0.69}\\
	  ELF w/o & \multirow{2}{*}{0.73} & \multirow{2}{*}{0.82} & \multirow{2}{*}{1.25} & \multirow{2}{*}{1.69} & \multirow{2}{*}{1.68} & \multirow{2}{*}{1.94} & \multirow{2}{*}{1.16} & \multirow{2}{*}{1.14} & \multirow{2}{*}{1.42}\\
	  pre-train &  &  &  &  &  &  &  &  & \\
	  Wi-Fi & 0.44 & 0.45 & 0.55 & 1.52 & 1.54 & 2.06 & 1.24 & 1.26 & 1.54\\
	  Geomag & 0.56 & 0.55 & 0.64 & 1.14 & 1.24 & 1.82 & 0.79 & 0.81 & 1.05\\
	  \hline
	\end{tabular}
	\begin{tablenotes}\footnotesize
        \item[1] Median error
        \item[2] Mean error
        \item[3] Third quartile of the error
    \end{tablenotes}
	\end{threeparttable}
\end{table}

\subsection{Trajectory Map Construction Performance}

Fig.~\ref{fig:map-floor-plan} shows the floor plans and ground-truth trajectories. Fig.~\ref{fig:map-IMU} shows the trajectories reconstructed via IMU-based dead reckoning, which deviate heavily from the ground truth due to the long-run drift problem. Fig.~\ref{fig:map-ELF}, ~\ref{fig:map-wifi}, and ~\ref{fig:map-geo} show the trajectories reconstructed by the SLAM systems using ELF, Wi-Fi RSSI, and geomagnetism. The trajectories reconstructed by ELF-SLAM are the closest to the ground truth. Table~\ref{tab:map-stats} lists the detailed mapping error statistics of the three modalities. 
ELF-SLAM achieves sub-meter mapping accuracy in all three environments, whereas Wi-Fi SLAM and geomagnetic SLAM's mapping errors increase in the large indoor space, i.e., office and mall. In \cite{lu2018simultaneous}, EMR SLAM using the smartphone earphone as the side-channel sensor yields about $1\,\text{m}$ to $2\,\text{m}$ median mapping errors in the evaluated office and lab spaces. Thus, ELF-SLAM outperforms Wi-Fi SLAM, geomagnetic SLAM, and EMR SLAM in map construction.

\subsection{Effectiveness of CL Pre-training}
\begin{figure}
    \centering
    \begin{minipage}{.45\columnwidth}
        \centering
    	\includegraphics[width=\columnwidth]{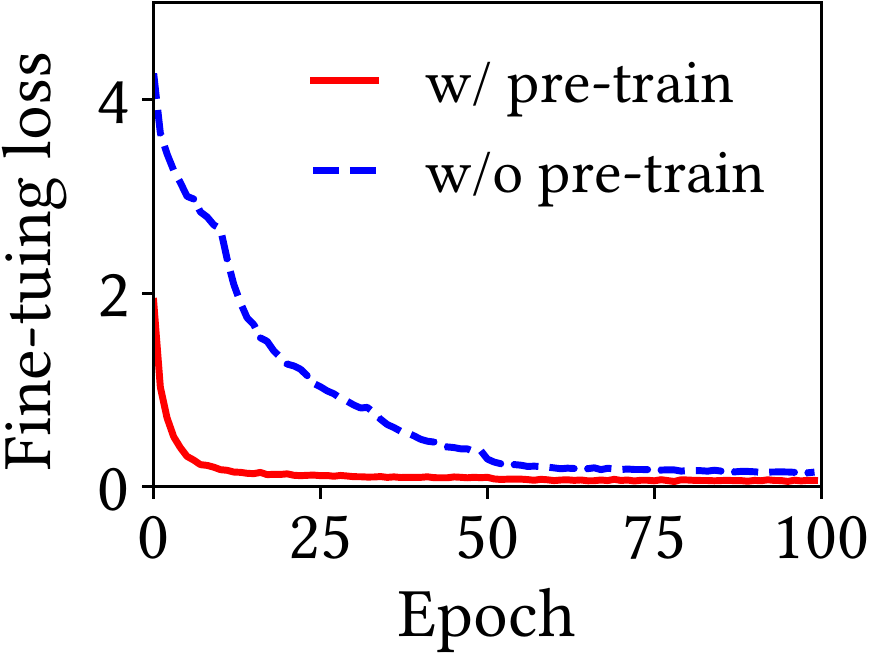}
		\vspace{-1em}
    	\caption{Trend of model finetuning loss.}
		\vspace{-0.5em}
    	\label{fig:loss-diff}
    \end{minipage}
	\hspace{0.5em}
	\vspace{-1em}
    \begin{minipage}{.47\columnwidth}
        \centering
		\includegraphics[width=\columnwidth]{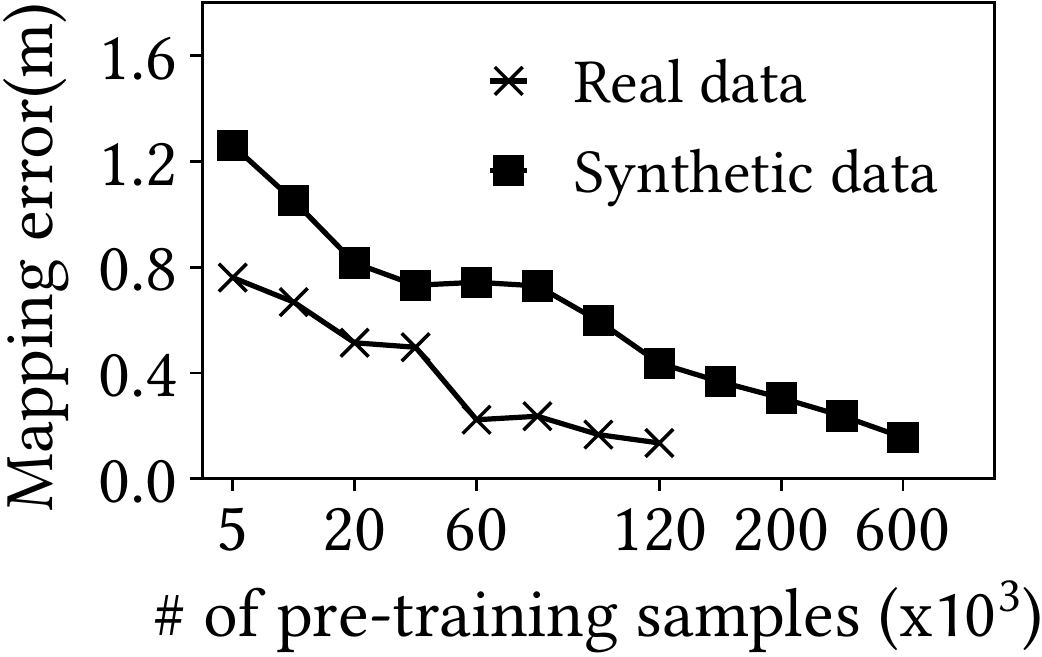} 
		\vspace{-1em}
		\caption{Impact of type and amount of pre-training data.}
		\vspace{-0.5em}
		\label{fig:eval-pretrain}    
    \end{minipage}
\end{figure}
We investigate the necessity of the CL pre-training by comparing the SLAM performance using the ELF extractors learned with and without the model pre-training step. The row ``ELF w/o pre-train'' in Table~\ref{tab:map-stats} is for the case without model pre-training. Compared with the result with model pre-training (row ``ELF''), the median mapping errors increase to $0.73\,\text{m}$, $1.16\,\text{m}$, and $1.69\,\text{m}$ in the three environments, respectively. Fig.~\ref{fig:loss-diff} shows the fine-tuning loss using the model with and without pre-training. The fine-tuning loss on the pre-trained model converges more quickly and is lower than that on the model without pre-training. The above results show that the model pre-training improves the efficiency of CL. 

We also investigate the model pre-training's requirement on data. We use two types of data for the pre-training, i.e., the synthetic data generated by the room acoustics simulator as described in \sect\ref{sec:loop-closure} and real data collected from the spaces different from the target space. For the latter, we use real echoes collected in the lab space, office, and shopping mall for model pre-training. Then, we perform fine-tuning and evaluation in the living room. Fig.~\ref{fig:eval-pretrain} shows the median mapping error versus the amount of pre-training data used. The median mapping errors decrease with pre-training data volume for both types of data. If real data is used for pre-training, the median mapping error converges to about $0.13\,\text{m}$ when data volume is more than $120\,\text{K}$. If synthetic data is used, the median mapping error converges to about $0.15\,\text{m}$ when data volume is more than $600\,\text{K}$. This shows that both the real data and synthetic data are useful for model pre-training. 
Although real data shows better efficiency in supporting the pre-training, synthetic data can be generated at massive scales and thus suffice for the pre-training.

\subsection{Impact of Finetuning Data Volume}

\begin{figure}
	\centering
	\includegraphics[width=0.55\columnwidth]{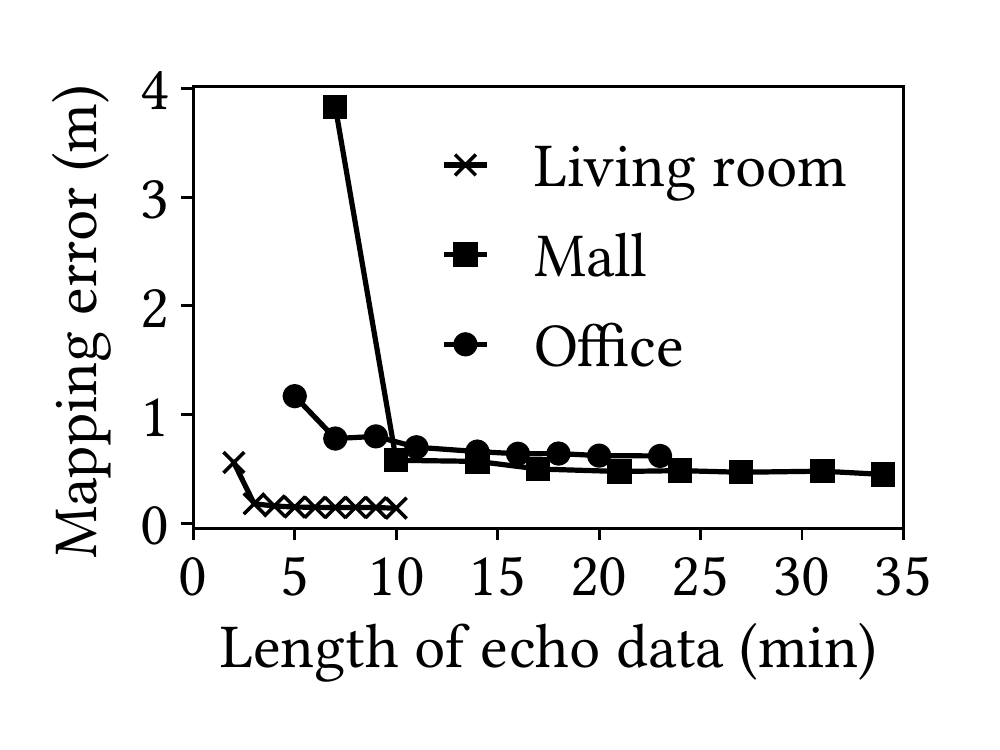}
	\caption{Impact of finetuning data volume on map construction performance.}
	\label{fig:eval-data-vol}
\end{figure}

We investigate the impact of the fintuning data volume on the performance of trajectory map construction. By default, we walk 4 to 5 rounds in the tested environments to collect the echoes. We gradually decrease the used data volume for model finetuning and show the corresponding trajectory map construction results in Fig.~\ref{fig:eval-data-vol}. In the living room, the median mapping error remains around $0.15\,\text{m}$ when the used data volume decreases from $10\,\text{min}$ to $3\,\text{min}$. The mapping error increases to $0.56\,\text{m}$ when the used data volume is $2\,\text{min}$. The increased error is due to no loop closure formed and the estimated trajectory cannot be corrected by the ELFs. Similarly, the mapping errors remain low when the used data volume is larger than $7\,\text{min}$ and $10\,\text{min}$ where valid loop closures can be detected via the ELFs in the office and the mall, respectively. The results show that the ELF-SLAM only requires a few minutes of data for model finetuning as long as there is loop closure formed during the walking.

\subsection{Localization Performance}
  
\begin{table}
	  \caption{Localization error statistics.}
	  \label{tab:loc-stats}
	  \begin{threeparttable}
	  \footnotesize
	  \centering
	  \begin{tabular}{c|ccc|ccc|ccc}
		\hline
		\multirow{2}{*}{Modality} & \multicolumn{3}{c|}{Living room} & \multicolumn{3}{c|}{Office} & \multicolumn{3}{c}{Mall}\\
		 & $\tilde{x}$ \tnote{1} & $\bar{x}$ \tnote{2} & Q3 \tnote{3} & $\tilde{x}$ & $\bar{x}$ & Q3 & $\tilde{x}$ & $\bar{x}$ & Q3 \\
		\hline
		\multicolumn{10}{c}{{\bf One-shot localization}}\\
		\hline
		{\bf ELF} & {\bf 0.10} & {\bf 0.29} & {\bf 0.14} & {\bf 0.54} & {\bf 0.60} & {\bf 0.80} & {\bf 0.42} & {\bf 0.79} & {\bf 0.67 }\\
		Wi-Fi & 1.67 & 2.17 & 3.30 & 3.44 & 4.27 & 6.16 & 3.04 & 3.86 & 5.22 \\
		Geomag & 1.06 & 2.31 & 3.93 & 2.19 & 3.95 & 5.38 & 12.48 & 13.36 & 19.28 \\
		\hline
		\multicolumn{10}{c}{{\bf Trajectory localization}}\\
		\hline
		{\bf ELF} & {\bf 0.10} & {\bf 0.22} & {\bf 0.64} & {\bf 0.41} & {\bf 0.54} & {\bf 0.97} & {\bf 0.47} & {\bf 0.53} & {\bf 0.86} \\
		WiFi & 0.54 & 0.730 & 1.13 & 1.74 & 1.86 & 3.09 & 1.46 & 1.90 & 3.73 \\
		Mag & 0.56 & 0.56 & 0.78 & 1.75 & 1.81 & 2.39 & 8.70 & 8.29 & 14.59 \\
		\hline
	  \end{tabular}
	  \begin{tablenotes}\footnotesize
		  \item[1] Median error
		  \item[2] Mean error
		  \item[3] Third quartile of the error
	  \end{tablenotes}
	  \end{threeparttable}
\end{table}

	  
We evaluate both the one-shot localization and trajectory localization of the three sensing modalities. Table~\ref{tab:loc-stats} lists the localization error statistics. 
For one-shot localization, ELF-SLAM achieves sub-meter median error in the three environments and outperforms both Wi-Fi SLAM and geomagnetic SLAM. 
For trajectory localization, each short trajectory consists of 8 consecutive footsteps. 
For Wi-Fi and geomagnetic SLAMs, the trajectory localization errors are less than the one-shot localization errors. For ELF-SLAM, trajectory localization does not bring much accuracy improvement over the one-shot localization, because the latter has already achieved a high localization accuracy close to the modality's spatial resolution. However, we will shortly show in \sect\ref{sec:sensitivity-analysis} that the trajectory localization brings performance improvement when ELF-SLAM is affected by various affecting factors.

\subsection{Sensitivity Analysis for Localization}
\label{sec:sensitivity-analysis}

\begin{figure}
    \centering
    \begin{minipage}{.475\columnwidth}
        \centering
    	\includegraphics[width=\columnwidth]{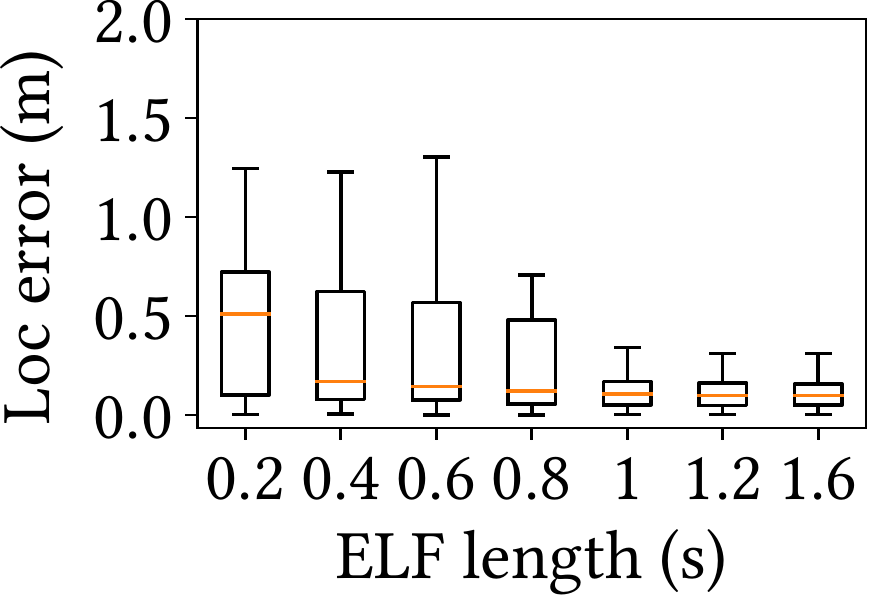}
		\vspace{-1em} 
    	\caption{ELF sequence length.}
    	\label{fig:eval-length}
		\vspace{-0.5em} 
    \end{minipage}
	\hspace{0.5em}
    \begin{minipage}{.475\columnwidth}
        \centering
		\includegraphics[width=\columnwidth]{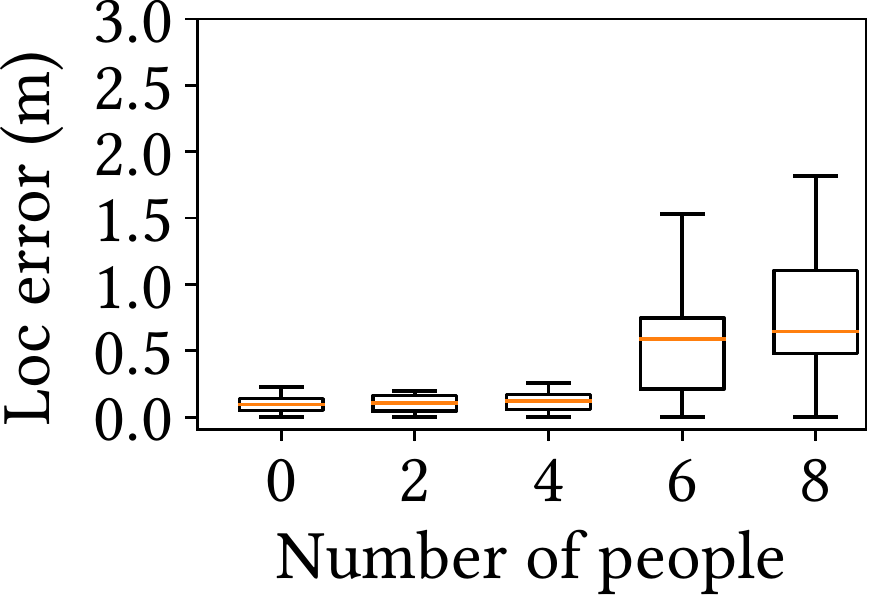} 
		\vspace{-2em} 
		\caption{Moving people.}
		\label{fig:eval-ppl}   
		\vspace{-0.5em} 
    \end{minipage}
\end{figure}

We conduct experiments mainly in the living room to evaluate the sensitivity of ELF-SLAM to various factors. By default, we consider one-shot localization.

\subsubsection{ELF sequence length.} 
\label{sec:elf-length}
Fig.~\ref{fig:eval-length} shows the localization errors when we vary the length of the ELF sequence used for computing ESS from $0.2\,\text{s}$ to $1.6\,\text{s}$. A boxplot shows the localization error distribution. The horizontal line in each boxplot shows the median. We can see that the localization error decreases with the ELF sequence length and becomes flat when the sequence length is more than $1\,\text{s}$. Note that at human's average walking speed, the duration between two consecutive footsteps is about $0.6\,\text{s}$, which results in an ELF sequence length of $0.6\,\text{s}$ as well. From Fig.~\ref{fig:eval-length}, at this length setting, the one-shot localization median error is around $0.1\,\text{m}$. Thus, ELF-SLAM performs well when the user walks at a normal speed.

\subsubsection{Nearby moving people.} Human bodies can reflect the excitation chirp and generate echoes irrelevant to ELF. Thus, we evaluate the impact of the nearby moving people on one-shot localization. We ask multiple volunteers to walk freely in the living room and talk to each other during the localization phase. Fig.~\ref{fig:eval-ppl} shows the localization error versus the number of nearby moving people. The localization error remains low when the number of people is up to 4. 
Note that the tested area is only about $60\,\text{m}^2$. When there are 6 and 8 moving people, whose crowd density is similar to that in the shopping mall during peak hours, the median localization errors increase to $0.57\,\text{m}$ and $0.6\,\text{m}$. Nevertheless, the errors remain at the sub-meter level. Thus, ELF-based localization can tolerate nearby moving people to a certain extent. 

\begin{figure}
    \centering
    \begin{minipage}{.47\columnwidth}
        \centering
		\includegraphics[width=\columnwidth]{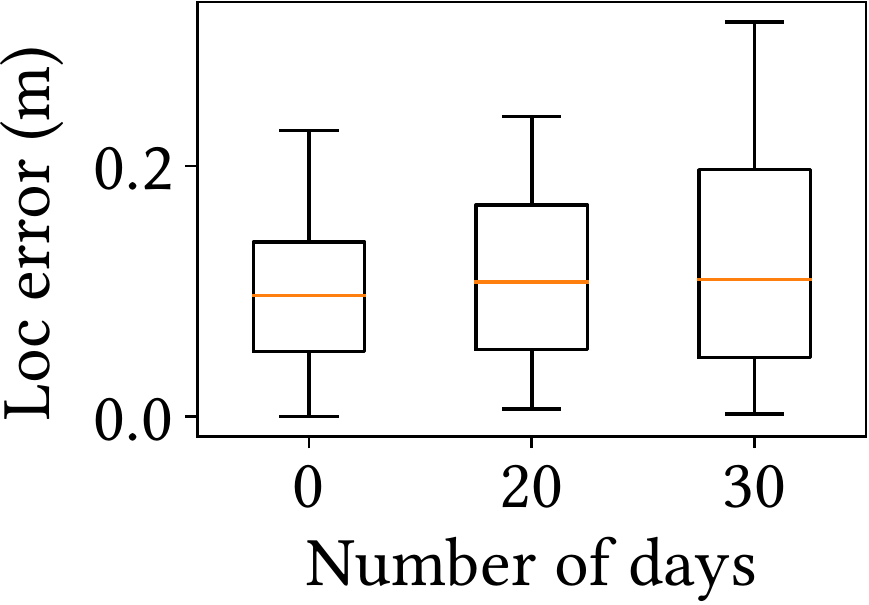} 
		\caption{Aging.}
		\label{fig:eval-aging}
    \end{minipage}
	\hspace{0.5em}
	\vspace{-1em}
    \begin{minipage}{.47\columnwidth}
        \centering
    	\includegraphics[width=\columnwidth]{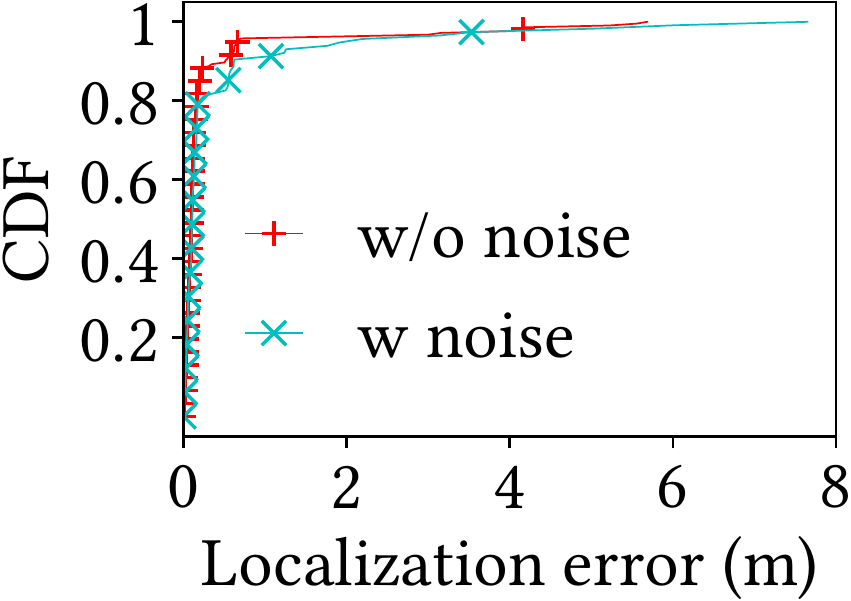} 
    	\caption{Audible noises.}
    	\label{fig:eval-noise} 
    \end{minipage}
\end{figure}


\begin{figure}[t]
	\begin{subfigure}[c]{.5\columnwidth}
		\includegraphics[width=\columnwidth]{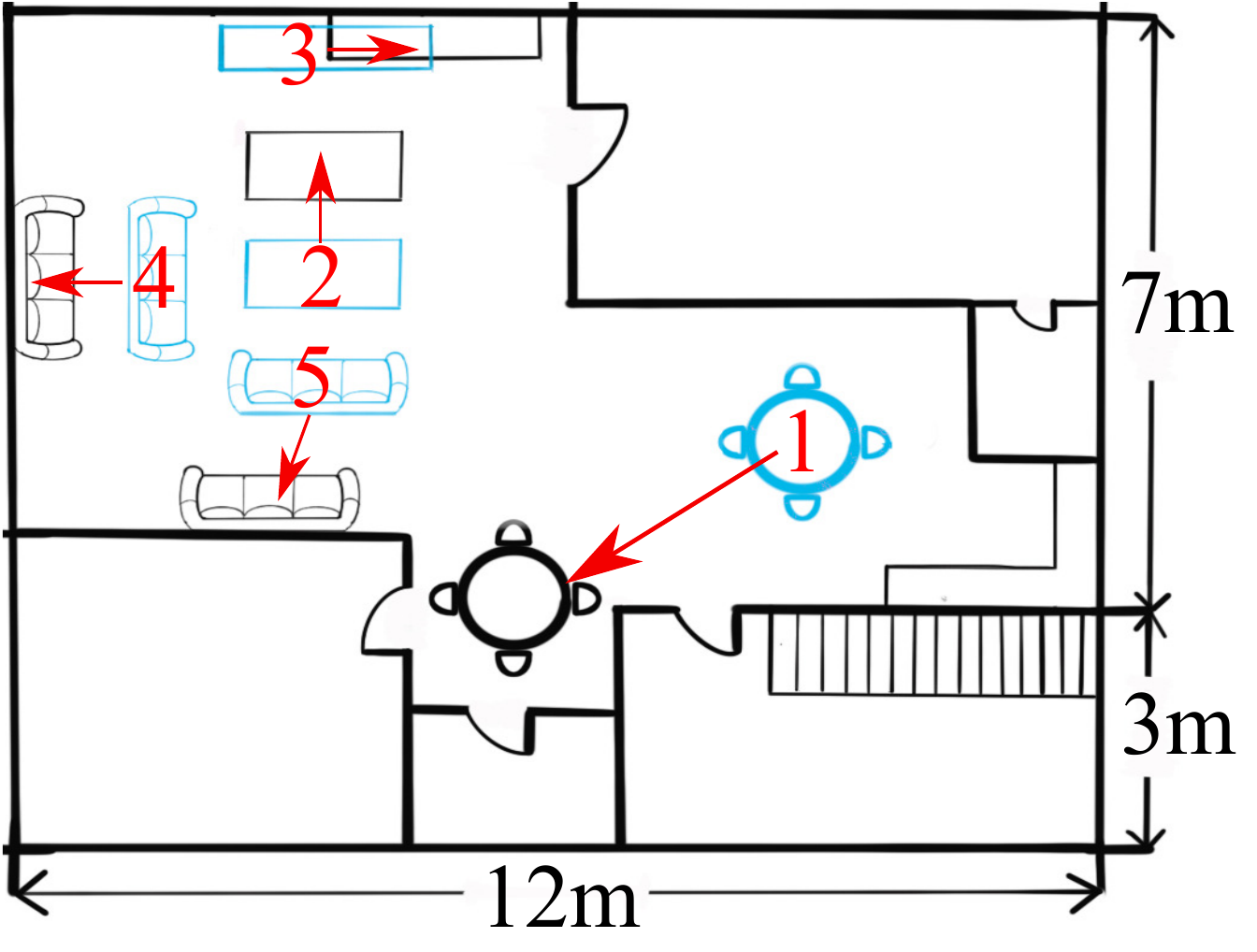}
		\caption{}
		\label{fig:furniture-1}
	\end{subfigure}
  \hfill
	\begin{subfigure}[c]{.45\columnwidth}
	    \includegraphics[width=\columnwidth]{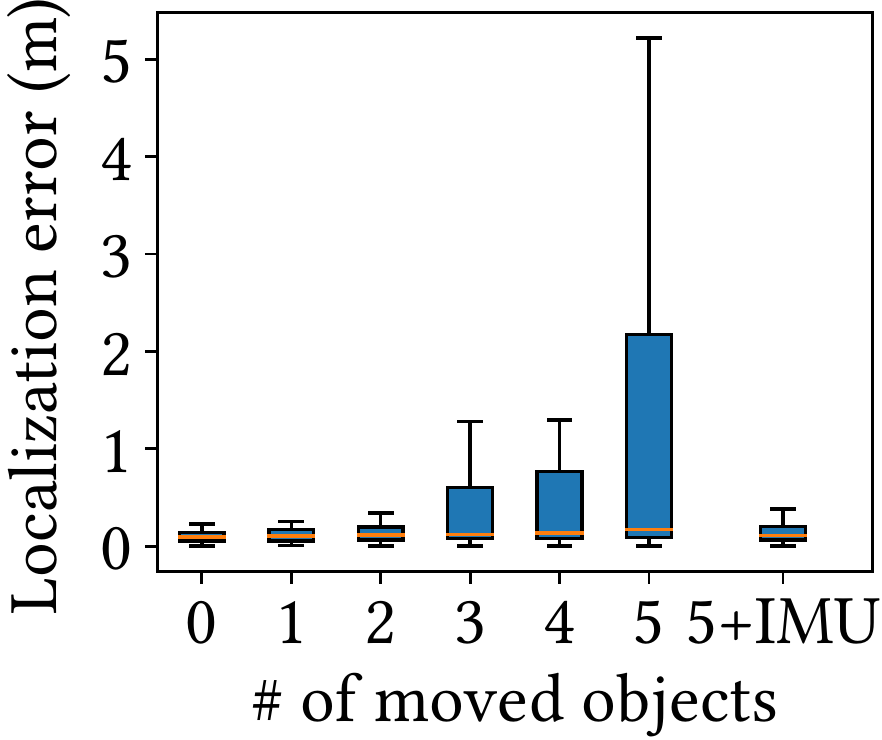} 
	    \caption{}
	    \label{fig:furniture-cdf}
	\end{subfigure}
	\caption{(a) Movements of furniture objects in a living room; (b) the corresponding localization performance.}
	\label{fig:eval-layout}
\end{figure}


\begin{figure}
    \centering
    \begin{minipage}{.43\columnwidth}
        \centering
        \includegraphics[width=\columnwidth]{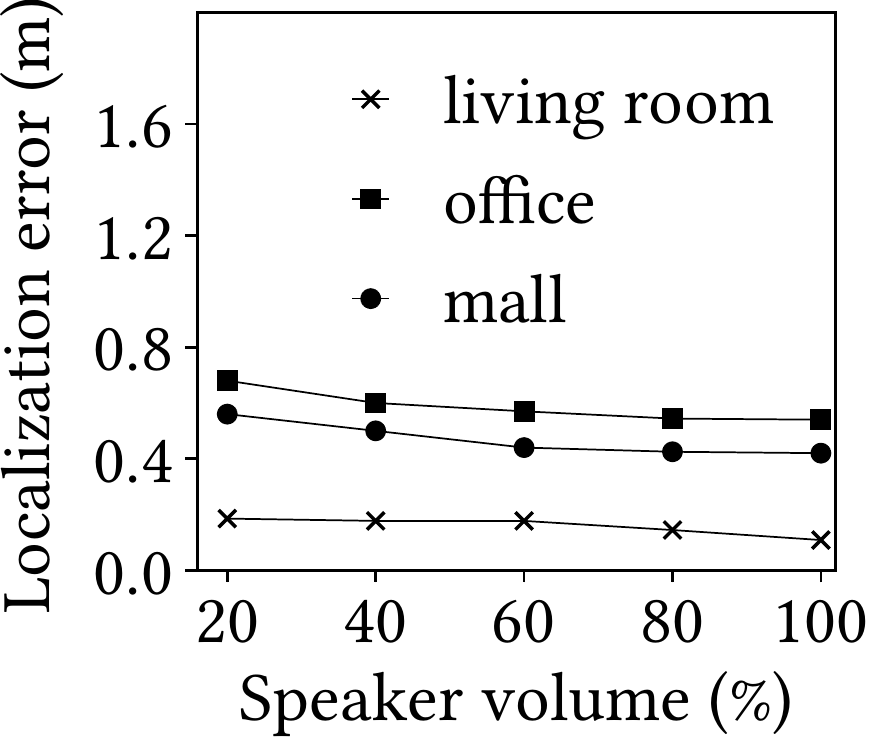}
		\vspace{-1em}
        \caption{Speaker volume.} 
		\vspace{-1em}
        \label{fig:eval-volume}   
    \end{minipage}
    \begin{minipage}{.55\columnwidth}
        \centering
		\includegraphics[width=\columnwidth]{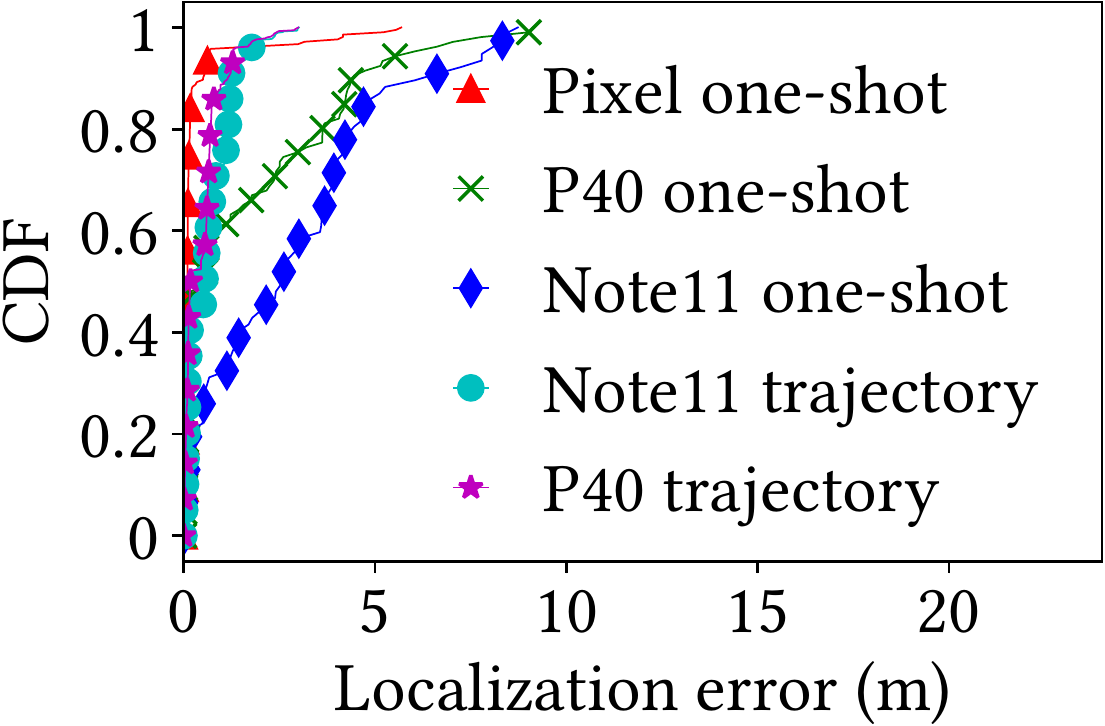} 
		\vspace{-1em}
		\caption{Hardware heterogeneity.}
		\vspace{-1em}
		\label{fig:eval-phone}
		\end{minipage}
\end{figure}



\subsubsection{Map aging.} 
We investigate whether the map constructed by ELF-SLAM ages. Specifically, at day 0, we use ELF-SLAM to construct a trajectory map. Then, we evaluate the ELF-based localization performance multiple times during a one-month period. Fig.~\ref{fig:eval-aging} shows the results. The median localization errors are $0.10\,\text{m}$, $0.11\,\text{m}$, and $0.12\,\text{m}$ at day 0, 20, and 30, respectively. This suggests that the constructed map does not have salient aging issue. In practice, a map can be continuously updated using the latest data contributed by users, to mitigate any potential aging issue.

\subsubsection{Audible noises.} We evaluate the robustness of ELF-based localization against audible noises. We use a laptop computer to play video clips of different contents (music, speech, etc) from Youtube to generate the noises. From Fig.~\ref{fig:eval-noise}, the noises have little impact on the localization performance. This is because our system operates within the near-inaudible frequency band. Thus, audible noises have negligible impact on ELF-based localization. 

\subsubsection{Space layout changes.} 
The layout changes of the target space may have impact on the chirp reverberation processes. Thus, we deliberately change the furniture locations in the living room to evaluate such impact. Fig.~\ref{fig:furniture-1} illustrates how the furniture objects are moved. 
Specifically, we move five objects including a dining table, a tea table, a TV cabinet, and two sofas. We move one object at a time. Fig.~\ref{fig:furniture-cdf} shows the localization error versus the number of moved objects. The error remains low when the number of moved objects is less than 5. When all the 5 objects are moved, the mean localization error increases to $1.3\,\text{m}$. If we apply the trajectory localization, the mean localization error decreases to $0.3\,\text{m}$ as labeled by ``5+IMU'' in Fig.~\ref{fig:furniture-cdf}.
Therefore, the trajectory localization improves the robustness of ELF-based localization against the layout changes. In practice, a map can be continuously updated using the latest data from the users to mitigate the impact of layout changes.

\subsubsection{Speaker volume.} 

As pets and human infants may have wider hearing limits \cite{hear-range}, they may perceive the chirps emitted from the smartphone. To avoid annoyance to them, we evaluate the localization with various settings for the smartphone's loudspeaker volume in emitting the chirps. Fig.~\ref{fig:eval-volume} shows the results. When the volume decreases from $100 \%$ (i.e., the highest volume) to $20 \%$, the median localization errors in the living room, office, and shopping mall increase from $0.1\,\text{m}$, $0.54\,\text{m}$, and $0.42\,\text{m}$ to $0.18\,\text{m}$, $0.68\,\text{m}$, and $0.56\,\text{m}$, respectively. Note that with 20\% loudspeaker volume, on the audible frequency band, the smartphone's sound is soft and becomes nearly imperceptible in environments with normal noise levels. Thus, ELF-SLAM maintains sub-meter accuracy when the chirp emission is soft. 

\subsubsection{Smartphone hardware heterogeneity.} 
\label{sec:hardware-hetero}
The microphone hardware heterogeneity can cause domain shifts for speech recognition \cite{mathur2018using}. To evaluate the impact of smartphone hardware heterogeneity on ELF-SLAM, we conduct experiments using three smartphones, i.e., Google Pixel 4, Huawei P40 Pro, and Redmi Note11. We use Pixel for map construction and all three smartphones for localization performance evaluation. Fig.~\ref{fig:eval-phone} shows the results. 
The median localization error in localizing Pixel is only $0.1\,\text{m}$. The median errors in localizing P40 and Note11 increase to $0.5\,\text{m}$ and $2.54\,\text{m}$, respectively. The hardware heterogeneity can be a primary reason for the performance drops, echoing the study \cite{mathur2018using}. Note that the Pixel and P40 have similar list prices, while the Note11 is about $3.5\,\times$ cheaper. This price comparison is consistent with the observation that Note11 experiences more performance drop than P40. 
We also evaluate the trajectory localization on the three smartphones. From the results labeled with ``trajectory'' in Fig.~\ref{fig:eval-phone}, for P40 and Note11, the median localization errors decrease to $0.2\,\text{m}$ and $0.5\,\text{m}$, respectively. Thus, the trajectory localization largely mitigates the negative impact of audio hardware heterogeneity. This result has the following two implications. First, inertial sensing, although suffering long-run drifts, provides important information for localization. Thus, fusing the results of inertial sensing and echo sensing increases the system's robustness. Second, because IMUs are in general low-cost, the three phones' IMUs may be of similar qualities. 

\subsection{Trajectory Map Superimposition}

\begin{figure}
    \centering
    \begin{minipage}{.485\columnwidth}
        \centering
        \includegraphics[width=\columnwidth]{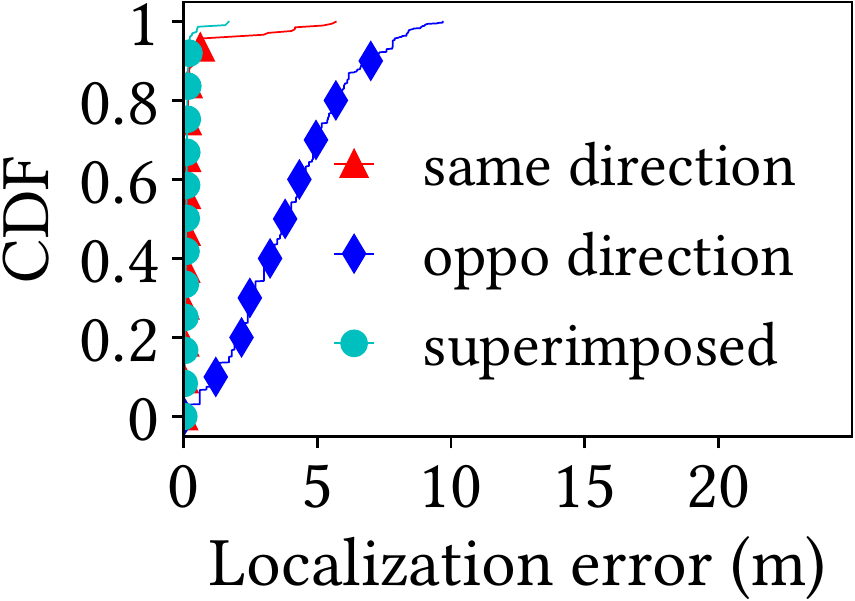}
		\vspace{-1em}
        \caption{Trajectory map superimposition on small-scale real echoes.} 
        \label{fig:eval_orient}   
    \end{minipage}
	\hspace{0.5em}
    \begin{minipage}{.475\columnwidth}
        \centering
	    \includegraphics[width=\columnwidth]{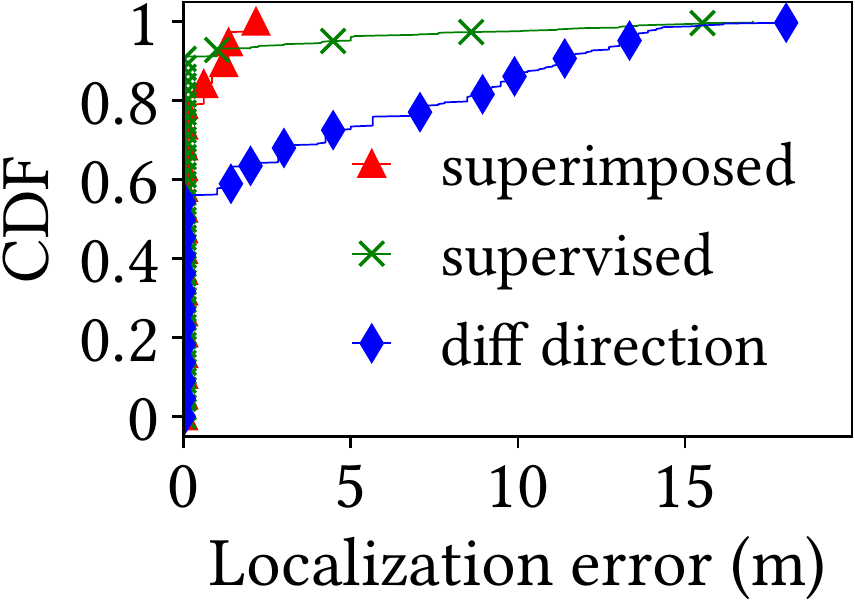} 
		\vspace{-1em}
	    \caption{Trajectory map superimposition on large-scale synthetic echoes.}
	    \label{fig:eval_orient_simul}
    \end{minipage}
\end{figure}

We evaluate the performance of the one-shot localization on the map superimposed from trajectory maps as described in \sect\ref{sec:superimpose}. 

\subsubsection{Evaluation on a small-scale dataset.} We conduct experiments in the living room. We follow the route in Fig.~\ref{fig:map-floor-plan} and walk in two opposite directions to generate two different trajectory maps. Then, we apply the proposed CL approach for map superimposition. Fig.~\ref{fig:eval_orient} shows the localization results. The plot labeled ``same direction'' is obtained when the smartphone during the localization phase is in the same orientation as the used map. The median localization error is $0.1\,\text{m}$. The curve labeled ``oppo direction'' is for the case when the smartphone during localization is in the opposite orientation as the trajectory map. The median localization error increases up to $3.8\,\text{m}$. The increased error is due to the sensitivity of ELF to large phone orientation deviations. 
The curve labeled ``superimposed'' shows the localization results using the map superimposed via the CL approach. The median localization error is $0.1\,\text{m}$, which is the same as the ``same direction'' result. This small-scale experiment shows that the map superimposed via CL can improve the ELF-based localization performance when trajectory maps with opposite directions are available.

\begin{figure}[t]
	\begin{subfigure}[c]{.55\columnwidth}
		\includegraphics[width=\columnwidth]{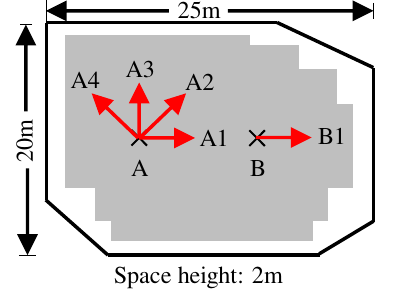}
		\caption{}
		\label{fig:room-simul}
		\end{subfigure}
  \hfill
	\begin{subfigure}[c]{.42\columnwidth}
	    \includegraphics[width=\columnwidth]{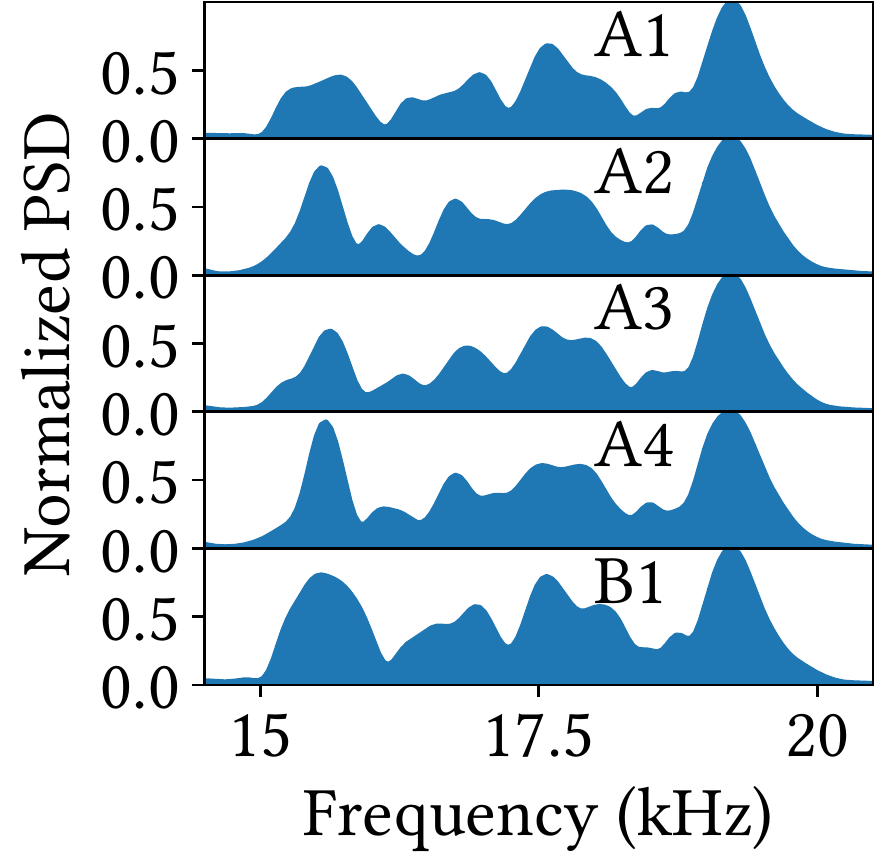} 
	    \caption{}
	    \label{fig:simul-psd}
	\end{subfigure}
	\caption{(a) Simulated space for extended evaluation of floor-level CL. (b) Spot A's echo PSDs on directions 1 to 4 and Spot B's echo PSD at direction 1.}
	\label{fig:simul-illustrate}
\end{figure}

\subsubsection{Evaluation on a large-scale synthetic dataset.} 
We also evaluate whether the floor-level CL can handle massive echo data from many trajectory maps. Note that this experiment omits trajectory map construction and only focuses on evaluating the superimposition performance, in terms of the one-shot localization error using the floor map. To allow the evaluation to easily scale up, we use the \texttt{pyroomacoustic} simulator to generate a large-scale synthetic dataset for an indoor space that has a polygon shape as shown in Fig.~\ref{fig:room-simul}.  
We collect data at 4,000 spots in the grey area where the distance between two neighbor spots is $10\,\text{cm}$. At each spot, we collect echo data when the simulated smartphone points to the directions as marked by the red arrows at spot A in Fig.~\ref{fig:room-simul}. For each orientation, we collect 100 echo traces with random perturbations to the orientation such that the traces are slightly different. As a result, a total of 16 million echo traces are collected. The first four rows of Fig.~\ref{fig:simul-psd} show the synthetic echoes' PSDs for four directions at spot A. They are slightly different from each other. The last row of Fig.~\ref{fig:simul-psd} shows the echo's PSD at spot B. It is different from all PSDs synthesized at spot A. This shows that the used simulator can synthesize both the orientation- and location-dependent echoes. Note that, from our estimation, collecting the same amount of data in the real world requires about 400 hours of manual labor. This experiment using synthetic data focuses on evaluating the scalability of the CL-based map superimposition algorithm. We leave the pilot deployment of our system that crowdsources many users' data for map superimposition to our future work. 

\begin{figure}[t]
	\begin{subfigure}[c]{.48\columnwidth}
		\includegraphics[width=\columnwidth]{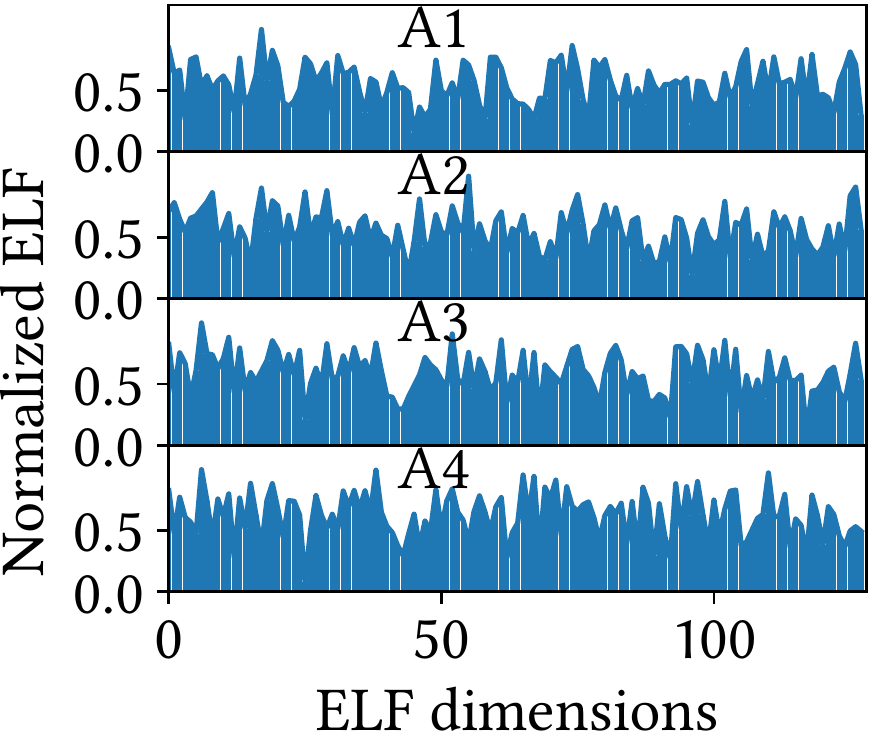}
		\caption{}
		\label{fig:floor-elf-wo-impose}
		\end{subfigure}
  \hfill
  \hspace{0.5em}
	\begin{subfigure}[c]{.48\columnwidth}
	    \includegraphics[width=\columnwidth]{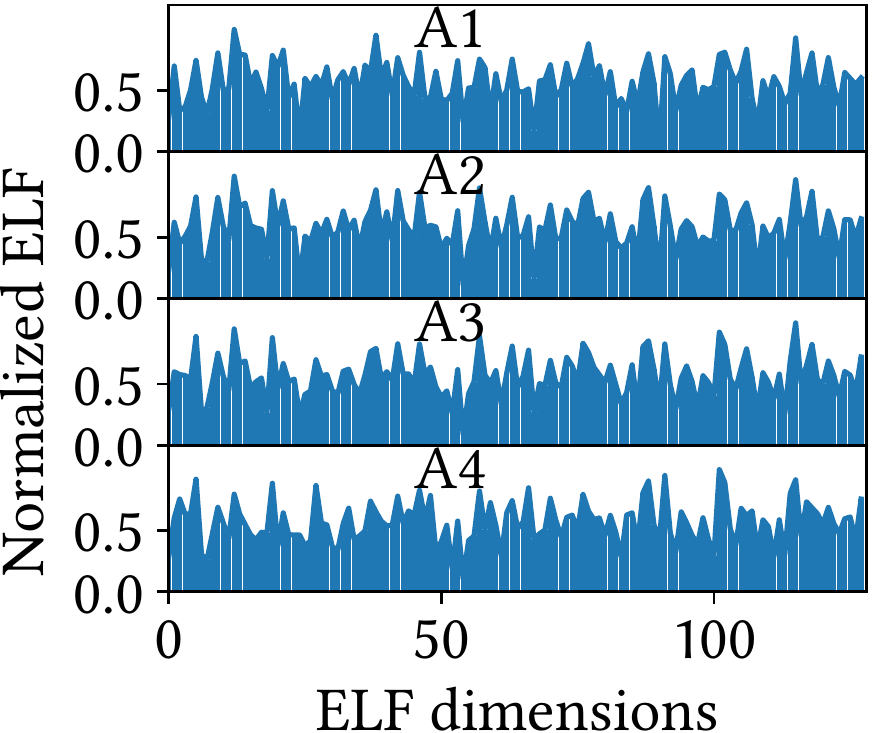} 
	    \caption{}
	    \label{fig:floor-elf-w-impose}
	\end{subfigure}
	\caption{Spot A's ELFs. (a) before map superimposition, the average similarity is 0.34; (b) after map superimposition, the average similarity is 0.76.}
	\label{fig:floor-elf}
\end{figure}

Fig.~\ref{fig:floor-elf} shows the spot A's ELFs from directions 1 to 4 before/after map superimposition. We calculate the cosine similarities among ELFs from directions 1 to 4. The ELFs' average similarity is 0.34 before the map superimposition. This value increases to 0.76 after applying the floor-level CL for map superimposition. This result shows that map superimposition is effective in reconciling the ELFs collected in different orientations. Fig.~\ref{fig:eval_orient_simul} shows the localization results on the synthetic data. The plot labeled ``diff direction'' shows the CDF when the CL-based map superimposition is not applied and the evaluated samples are in a different phone orientation from that in the used map. The mean localization error is $3.2\,\text{m}$. This poor result shows the necessity of the CL-based reconciliation. The curve labeled ``superimposed'' shows the results obtained using the floor map constructed by the floor-level CL. The mean localization error decreases to $0.24\,\text{m}$. We also employ the supervised fingerprint approach as a baseline, which forms the training dataset by labeling the echoes synthesized at the same spot with the same location label and trains a DNN to classify the 4,000 spots. The CDF curve labeled ``supervised'' shows the results. The mean localization error is $0.56\,\text{m}$. The supervised fingerprint approach is inferior to the proposed solution that performs localization using the floor map.


\subsection{System Overhead}

\begin{figure}
    \centering
    \begin{minipage}{.9\columnwidth}
        \centering
	    \includegraphics[width=\columnwidth]{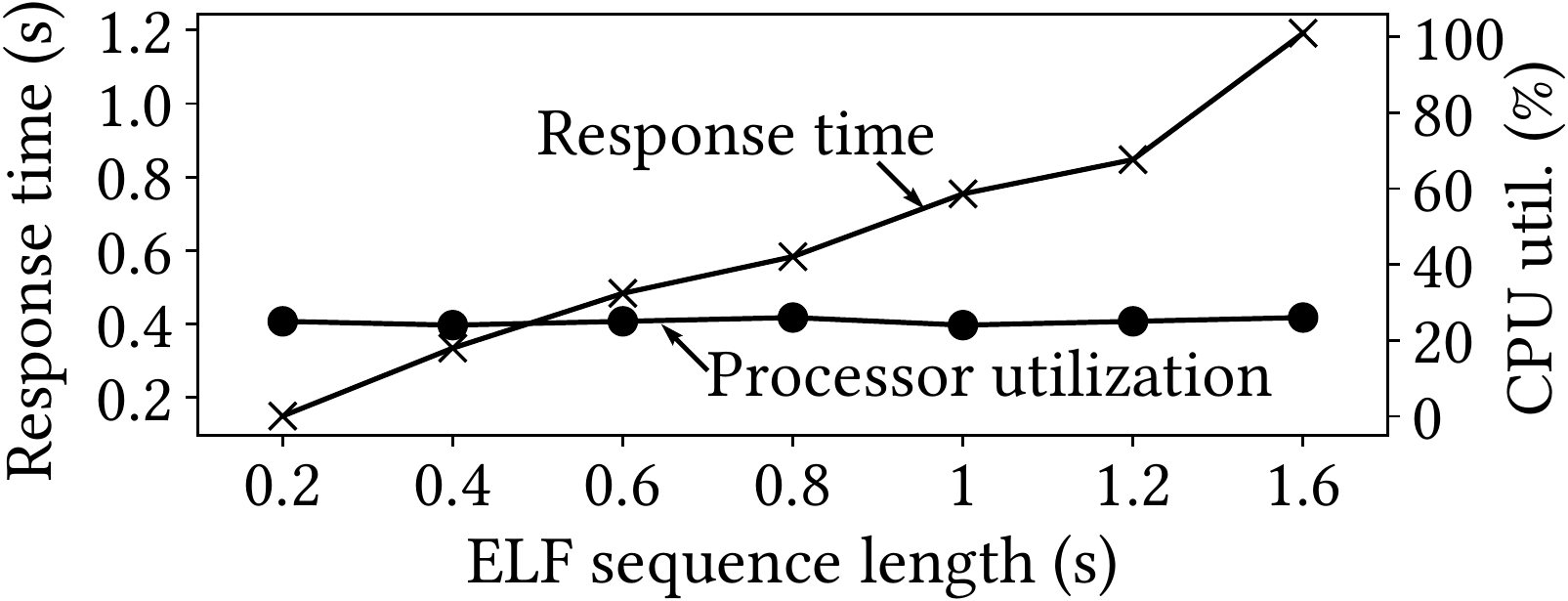} 
	    \caption{Model execution overhead.}
	    \label{fig:eval-overhead}
    \end{minipage}
\end{figure}

We evaluate the computation overheads of the ELF extractor yielded by the floor-level CL and real-time one-shot localization using the floor map on the Google Pixel 4 smartphone. We use \texttt{Pytorch-Mobile} \cite{pytorch-mobile} to optimize and deploy the ELF extractor model. The compressed model is about $96\,\text{MB}$. 

\subsubsection{App's response time and processor utilization.} When the ELF sequence length varies from $0.2\,\text{s}$ to $1.6\,\text{s}$, from Fig.~\ref{fig:eval-overhead}, the smartphone processor utilization remains at around 20\%. The memory usage of storing $4,000$ spots' ELFs is less than $4\,\text{MB}$. We also measure the app's response time, which includes the times for extracting ELFs and matching the ELF trace against the floor map.
The response time increases from $0.18\,\text{s}$ to $1.2\,\text{s}$, when the ELF sequence length varies from $0.2\,\text{s}$ to $1.6\,\text{s}$. The increased response time is from the localization module, because the computation overhead of the feature matching increases with the ELF sequence length. From Fig.~\ref{fig:eval-length}, by setting the ELF sequence length to be $0.6\,\text{s}$, our system achieves $0.1\,\text{m}$ median localization error, while the corresponding measured response time is about $0.5\,\text{s}$. Thus, the user can get the localization result in about $1.1\,\text{s}$.

\subsubsection{App's network bandwidth and battery usages.} To continuously transmit echo and IMU data to the cloud server for map construction, the app's bandwidth usage is around $90\,\text{kbps}$. This data rate is similar to that of Advanced Audio Coding (AAC), a widely adopted standard for lossy audio compression. Note that as the localization phase of ELF-SLAM is performed locally on the phone, it requires no data transmission. We use the \texttt{battery historian}  \cite{battery-hostorian} to estimate the app's energy usage. The app's energy usage per hour is around $270\,\text{mAh}$ when the app performs localization continuously. This energy usage is similar to that of the Google Map app in continuous navigation, i.e., around $280\,\text{mAh}$ and much lower than a visual SLAM \cite{vi-slam}, whose energy consumption is around $450\,\text{mAh}$. Thus, our ELF-based localization system introduces acceptable overhead. 


\section{Discussion}
\label{sec:discussion}

$\blacksquare$ {\bf Concurrent use and security.} It is common to have multiple users simultaneously use their smartphones for location sensing in same indoor space. To avoid signal collision, we will implement the carrier-sense multiple access (CSMA) protocol in the ELF-SLAM system to manage the traffic. Specifically, when a smartphone is about to transmit the chirp, it uses its microphone to detect whether there is an ongoing echoing process. If so, the smartphone will defer the chirp transmission. The ELF-SLAM can be vulnerable to malicious attacks. For example, attackers can deploy speakers in an indoor space and play the sound in the used band to mislead the system. The development of the defense mechanism will be considered in future work. 

$\blacksquare$ {\bf Domain adaptation across different smartphones.} \sect\ref{sec:hardware-hetero} shows the impact of smartphone heterogeneity on the ELF-SLAM. We have applied {\em trajectory localization} to improve the ELF-SLAM's performance across different smartphones. Several machine learning techniques are also available to address the domain shift problem, e.g., data augmentation \cite{mathur2018using}, few-shot learning \cite{gong2019metasense}, and adversarial learning \cite{mathur2019mic2mic}, etc. We will continue to explore a more effective way to address smartphone heterogeneity in future work.

$\blacksquare$ {\bf Impact of inaudible sound to human.} 
The experimental results in \cite{van2014undesirable} show that participants experience mild side effect after 20 minutes' exposures in frequency from $12.5\,\text{Hz}$ to $20\,\text{kHz}$ with sound level from 44 to $71\,\text{dB}$. We investigate the played chirp level on Google Pixel 4 by increasing the sound volume from 20\% to 100\%, and find that the corresponding sound level is between 40 - $68\,\text{dB}$. As shown in Fig.~\ref{fig:eval-data-vol}, ELF-SLAM requires less than $20\,\text{min}$ data collection in the tested environment. Thus, ELF-SLAM will have little impact on humans within tens of minutes exposure.


\section{Conclusion}
\label{sec:conclude}

This paper presents ELF-SLAM, an indoor smartphone SLAM system using learning-based ELFs. ELF-SLAM uses a smartphone's built-in audio hardware to emit near-inaudible chirps and record acoustic echoes in an indoor space, then uses the echoes to detect loop closures for regulating the IMU-based dead reckoning. To effectively capture loop closures, we design a trajectory-level contrastive learning procedure and apply it on the echoes to learn ELFs. Then, we design a clustering-based approach to remove the false detection results and curate the loop closures. Lastly, we design floor-level contrastive learning to superimpose the trajectory maps such that the differences among the ELFs collected in different phone orientations are reconciled. Our extensive experiments show that ELF-SLAM achieves sub-meter accuracy in both mapping and localization, and outperforms Wi-Fi RSSI and geomagnetic SLAMs. 




\begin{acks}
This study is supported under the RIE2020 Industry Alignment Fund – Industry Collaboration Projects (IAF-ICP) Funding Initiative, as well as cash and in-kind contribution from Singapore Telecommunications Limited (Singtel), through Singtel Cognitive and Artificial Intelligence Lab for Enterprises (SCALE@NTU).
\end{acks}


\bibliographystyle{ACM-Reference-Format}
\bibliography{sample-base}


\end{document}